\documentclass{article}

\usepackage{authblk}
\usepackage{fullpage}

\usepackage{microtype}
\usepackage{graphicx}
\usepackage{subfig}
\usepackage{tabu}
\usepackage{booktabs} 
\usepackage{enumitem}
\usepackage{amsmath, amsthm}
\usepackage[numbers,sort&compress,square]{natbib}
\usepackage{algorithm, algorithmic}

\usepackage{hyperref}
 \usepackage{comment}
 \usepackage[utf8]{inputenc}
\usepackage[english]{babel}

\usepackage{Definitions}

\newcommand{\UnnumberedFootnote}[1]{{\def\thefootnote{}\footnote{#1}
\addtocounter{footnote}{-1}}}

\newcommand{\xhdr}[1]{\vspace{1mm} \noindent{{\bf #1.}}}

\newcommand{\given}{{\,|\,}}

\graphicspath{{./FIG/}}

\begin{document}

\title{Optimal Decision Making Under Strategic Behavior}

\author{Stratis Tsirtsis$^{\dagger}$$^{*}$}
\author{Behzad Tabibian$^{\mathsection}$$^{*}$}
\author{Moein Khajehnejad$^{\ddag}$$^{*}$}           
\author{Adish Singla$^{\dagger}$}
\author{Bernhard~Sch\"{o}lkopf$^{\mathparagraph}$}
\author{Manuel Gomez Rodriguez$^{\dagger}$}
\affil{$^{\dagger}$MPI for Software Systems, \{stsirtsis, adishs, manuelgr\}@mpi-sws.org \\
$^{\mathsection}$Reasonal, Inc., behzad@reason.al \\
$^{\ddag}$Monash University, moein.khajehnejad@monash.edu \\
$^{\mathparagraph}$MPI for Intelligent Systems, bs@tuebingen.mpg.de}

\date{}

\clubpenalty=10000
\widowpenalty = 10000

\maketitle

\UnnumberedFootnote{\scriptsize $^{*}$Authors contributed equally. Moein Khajehnejad contributed to this work during his internship at the Max Planck Institute for Software 
Systems. Behzad Tabibian contributed to this work while he was a graduate student at the Max Planck Institute for Intelligent Systems and the Max Planck Institute 
for Software Systems.}

\begin{abstract}
We are witnessing an increasing use of data-driven predictive mo\-dels to inform decisions. As decisions have 
implications for in\-di\-vi\-duals and society, there is increasing pressure on decision ma\-kers to be transparent 
about their decision policies. 
At the same time, individuals may use know\-led\-ge, gained by transparency, to invest effort strategically in 
order to maximize their chances of receiving a beneficial decision.
Our goal is to find decision policies that are optimal in terms of uti\-li\-ty in such a strategic setting.
To this end, we first characterize how strategic investment of \-effort by individuals leads to a change in the feature 
distribution.
Using this characterization, we first show that, in general, we cannot expect to find op\-timal decision policies in 
polynomial time and there are cases in which deterministic policies are suboptimal. 
Then, we demonstrate that, if the cost individuals pay to change their features satisfies a natural mo\-no\-to\-ni\-ci\-ty 
assum\-ption, 
%
%
we can narrow down the search for the optimal policy to a particular family of decision policies with a set of desirable
properties, which allow for a highly effective polynomial time heuristic search algorithm using dynamic programming.
%
%
Finally, under no assump\-tions on the cost individuals pay to change their features, we develop an iterative search 
algorithm that is guaranteed to find locally optimal decision policies also in polynomial time.
Experiments on synthetic and real credit card data illus\-tra\-te our theoretical findings and show that the decision policies found by our 
algorithms achieve higher utility than those that do not account for strategic behavior.
\end{abstract}


\vspace{-3mm}
\section{Introduction}
\label{sec:introduction}
\vspace{-1mm}
Consequential decisions across a wide variety of domains, from banking and hiring to insurances, are increasingly informed by data-driven 
predictive models.
In all these domains, the decision maker aims to employ a decision policy that maximizes a given utility function while the predictive model aims to provide an 
accurate prediction of the outcome of the process from a set of observable features.
For example, in loan decisions, a bank may decide whether or not to offer a loan to an applicant on the basis of a predictive model's estimate of the probability that 
the individual would repay the loan.

In this context, there is an increasing pressure on the decision makers to be transparent about the decision policies, the predictive models, 
and the features they use. 
However, individuals are in\-cen\-ti\-vized to use this knowledge to invest effort strategically in order to receive a beneficial decision.
With this motivation, there has been a recent flurry of work on strategic classification~\cite{bruckner2012static, bruckner2011stackelberg, dalvi2004adversarial, dong2018strategic, hardt2016strategic, hu2019disparate, milli2018social, miller2019strategic, frankel2019improving, braverman2020role}. 
This line of work has focused on developing accurate predictive models and it has shown that, under certain technical conditions, it is possible 
to protect predictive models against misclassification errors that would have resulted from this strategic behavior. 
In this work, rather than accurate predictive models, we pursue the development of decision policies that maximize utility in this strategic
setting.
Our work is most closely related to a recent line of work on incentive-aware evaluation mechanisms~\cite{kleinberg2018classifiers, alon2020multiagent}, which aims to design scoring 
rules that incentivize agents to invest effort in specific actions. 
However, in these works, the decision maker does not employ a predictive model and the feature distribution of a population of agents to design her decision policy. As a result, the technical contributions of these works are orthogonal to ours.
More broadly, our work also relates to recent work on the long-term consequences of machine learning algorithms~\cite{hu2018short, liu2018delayed, mouzannar2019fair, Tabibian2019}, recommender systems~\cite{schnabel2018short, sinha2016deconvolving} and counterfactual explanations~\cite{wachter2017counterfactual, ustun2019actionable, tsirtsis2020decisions}.

Once we focus on the utility of a decision policy, it is overly pessimistic to always view an individual's strategic effort as some form of gaming, 
and thus undesirable---an individual's effort in changing their features may actually lead sometimes to self-improvement, as noted by several 
studies in economics~\cite{coate1993will, fryer2013valuing, hu2018short} and, more recently, in the theoretical computer science 
literature~\cite{kleinberg2018classifiers, alon2020multiagent, haghtalab2020}.
For example, in car insurance decisions, if an insurance company uses the number of speeding tickets a driver receives to decide how much to charge 
the driver, she may feel compelled to drive more carefully to pay a lower price, and this will likely make her a better driver.
In loan decisions, if a bank uses credit card debt to decide about the interest rate it offers to a customer, she may feel compelled to avoid overall credit card
debt to pay less interest, and this will improve her financial situation.
In hiring decisions, if a law firm uses the number of internships to decide whether to offer a job to an applicant, she may feel compelled to do more 
internships during her studies to increase their chances of getting hired, and this will improve her job performance.
In all these scenarios, the decision maker---insurance company, bank, or law firm---would like to find a decision policy that incentivizes individuals to 
invest in forms of effort that increase the utility of the policy---reduce payouts or default rates, or increase job performance.

In this work, we cast the problem as a Stackelberg game in which the decision maker moves first and shares her decision policy before individuals 
best-respond and invest effort to maximize their chances of receiving a beneficial decision under the policy.
Here, we assume that the decision maker takes decisions based on low dimensional feature vectors since, in many realistic scenarios, the 
data is summarized by just a small number of summary statistics (\eg, FICO scores)~\cite{hardt2016equality,liu2018delayed}.
Then, we characterize how this strategic investment of effort leads to a change in the feature distribution at a population level.
More specifically,~we~de\-rive an analytical expression for the feature distribution induced by \emph{any} policy in terms 
of the original feature distribution by solving an optimal transport problem~\cite{villani2008optimal}.
Based on this analytical expression, we make the following contributions:
\begin{itemize}[leftmargin=0.7cm] 
\item[I.] We show that the problem of finding the optimal decision policy is NP-hard by using a novel reduction of the Boolean satisfiability (SAT) problem~\cite{karp1972reducibility}. 

\item[II.] We show that there are cases in which deterministic policies are suboptimal in terms of utility. This is in contrast 
with the non-strategic setting, where de\-ter\-mi\-nis\-tic threshold rules are optimal~\cite{corbett2017algorithmic, valera2018enhancing}.

\item[III.] Under a natural monotonicity assumption on the cost individuals pay to change features~\cite{hardt2016strategic, hu2019disparate}, we 
show that 
we can narrow down the search for the optimal policy to a particular family of decision policies with a set of desirable properties.
Moreover, these properties allow for the development of a highly effective polynomial time heuristic search algorithm using dynamic 
programming (refer to Algorithm~\ref{alg:dp}).
%

\item[IV.] Under no assumptions on the cost individuals pay to change features, we introduce an iterative search algorithm that is guaranteed
to find locally optimal decision policies also in polynomial time  (refer to Algorithm~\ref{alg:iterative})
\end{itemize}
%
%

Finally, we expe\-riment with synthetic and real credit card data to illustrate our theoretical findings and show that the decision policies 
found by our algorithms achieve higher utility than several competitive baselines\footnote{\scriptsize To facilitate research in this area, we release
an open-source implementation of our algorithms at \href{https://github.com/Networks-Learning/strategic-decisions}{https://github.com/Networks-Learning/strategic-decisions}.}.
 
%


\vspace{-2mm}
\section{Decision policies, utilities, and benefits}
\label{sec:preliminaries}
\vspace{-1mm}

%
Given an individual with a feature vector $\xb \in \{1, \ldots, n\}^{d}$ and a (\emph{ground-truth}) label $y \in \{0, 1\}$, a decision $d(\xb) \in \{0, 1\}$ controls whether the 
label $y$ is \emph{realized}\footnote{\scriptsize For simplicity, we assume features are discrete and, without loss of generality, we assume each feature takes $n$ discrete values.} .
This setting fits a variety of real-world scenarios, where continuous features are often discretized into (percentile) ranges.
As an example, in a loan decision, the decision specifies whether the individual receives a loan ($d(\xb) = 1$) or her application is rejected 
($d(\xb) = 0$); the label indicates whether an individual repays the loan ($y = 1$) or defaults ($y = 0$) upon receiving it; and the feature vector 
($\xb$) may include an individual'{}s salary percentile, education, or credit history.
Moreover, we denote the number of feature values using $m = n^d$, assuming that the number of features $d$ is small as discussed in Section~\ref{sec:introduction}.

Each decision $d(\xb)$ is sampled from a decision policy $d(\xb) \sim \pi(d \given \xb)$ and, for each individual, the labels $y$ are sampled from $P(y \given \xb)$.
Throughout the paper, for brevity, we will write $\pi(\xb) = \pi(d=1 \given \xb)$ and we will say that the decision policy satisfies \emph{outcome monotonicity} if the 
higher an individual'{}s outcome, the higher their chances of receiving a positive decision, \ie, 
\begin{equation} \label{eq:outcome-monotonic-policy}
P(y=1 \given \xb_i) < P(y=1 \given \xb_j) \Leftrightarrow \pi(\xb_i) < \pi(\xb_j)
\end{equation}
%
%
Moreover, we adopt a Stackelberg game-theoretic formulation in which the decision maker publishes her decision policy $\pi$ before individuals (best-)respond. 
As it will become clearer in the next section, individual best responses lead to a change in the feature distribution at a population level---we will say that the new 
feature distribution $P(\xb \given \pi)$ is induced by the policy $\pi$. 
Then, we measure the (immediate) utility a decision maker obtains using a policy $\pi$ as the average overall profit she obtains~\cite{corbett2017algorithmic, valera2018enhancing, kilbertus2019improving}, \ie,
\begin{align} \label{eq:utility}
u(\pi, \gamma) &= \EE_{\xb \sim P(\xb \given \pi),\, y\sim P(y\given\xb),\, d \sim \pi(d \given \xb)} [y~d(\xb) - \gamma~d(\xb)] \nonumber \\  
&= \EE_{\xb\sim P(\xb\given\pi),\, d \sim \pi(d \given \xb)} [P(y=1 \given \xb )~d(\xb) - \gamma~d(\xb)].
\end{align}
where $\gamma \in (0, 1)$ is a given constant reflecting economic considerations of the decision maker. For example, in a loan scenario, the term $d(\xb) P(y=1 \given \xb )$ is proportional to the expected number of individuals who receive and repay a loan, the term $d(\xb) \gamma$ is proportional to the number of individuals who receive a loan, and $\gamma$ measures the cost of offering a loan in units of repaid loans.
Here, note that $\gamma$ is bounded by the collateral against the loan, which caps the maximum potential cost to the loan provider.
Finally, we define the (immediate) individual benefit an individual with features $\xb$ obtains from a policy $\pi$ as
\begin{equation} \label{eq:benefit}
b(\xb) = \EE_{d \sim \pi(d \given \xb)} [f(d(\xb))], 
\end{equation}
where the function $f(\cdot)$ is problem dependent. 
Here, for ease of exposition, we will assume that $f(d(\xb)) = d(\xb)$ and thus 
$b(\xb) = \EE_{d \sim \pi(d \given \xb)}[d(\xb)] = \pi(\xb)$, however, our results can be extended to any function $f(\cdot)$ that is proportional to the probability that
individuals receive a positive decision.
%
%

\xhdr{Remarks on strategic classification} 
Due to Goodhart'{}s law, if $\xb$ are noncausal, then $\xb$ can lose predictive power for $y$ after individuals (best-)respond, \ie, $P(y \given \xb)$
may change. This has been a key insight by previous work on strategic classification~\cite{hardt2016strategic, hu2019disparate, milli2018social, shavit2020causal, bechavod2020causal}, which 
aims to develop accurate predictive models $P_{\theta}(y \given \xb)$ in a strategic setting. 
Even if $\xb$ are causal and $P(y \given \xb)$ does not change after best-response, a predictive model $P_{\theta}(y \given \xb)$ trained using empirical 
risk minimization, \ie, $\theta^{*} = \argmin_{\theta} \EE_{\xb \sim P(\xb),\, y\sim P(y\given\xb)}[\ell(\xb, y, \theta)]$, where $\ell(\cdot)$ is a given loss function, may decrease its accuracy after best-response. 
This is because, once individuals best respond to a decision policy $\pi$, the parameters $\theta^{*}$ may be suboptimal with respect to the feature distribution induced 
by the policy, \ie, 
\begin{equation*}
\EE_{\xb, y \sim P(\xb \given \pi) P(y\given\xb)}[\ell(\xb, y, \theta^{*})] \geq \min_{\theta} \EE_{\xb, y \sim P(\xb \given \pi) P(y\given\xb)}[\ell(\xb, y, \theta)].
\end{equation*}
In the context of strategic classification,~\citet{miller2019strategic} have very recently argued that (best-)responses to noncausal and causal features correspond to gaming and improvement, respectively.
In this work, for simplicity, we assume that $P(y \given \xb)$ does not change and $P_{\theta}(y \given \xb) = P(y \given \xb)$, however, the development of optimal policies 
that account for changes on $P(\xb)$, $P(y \given \xb)$ and $P_{\theta}(y \given \xb)$ after individuals best-respond is a very interesting direction for future 
work~\cite{miller2019strategic,shavit2020causal}.
%

\vspace{-2mm}
\section{Problem Formulation}
\label{sec:formulation}
\vspace{-1mm}
\begin{figure*}[t]
     \centering
     \begin{tabu} to 1\textwidth { X[c]  X[c]  X[c] X[c] }
       \includegraphics[scale=0.25]{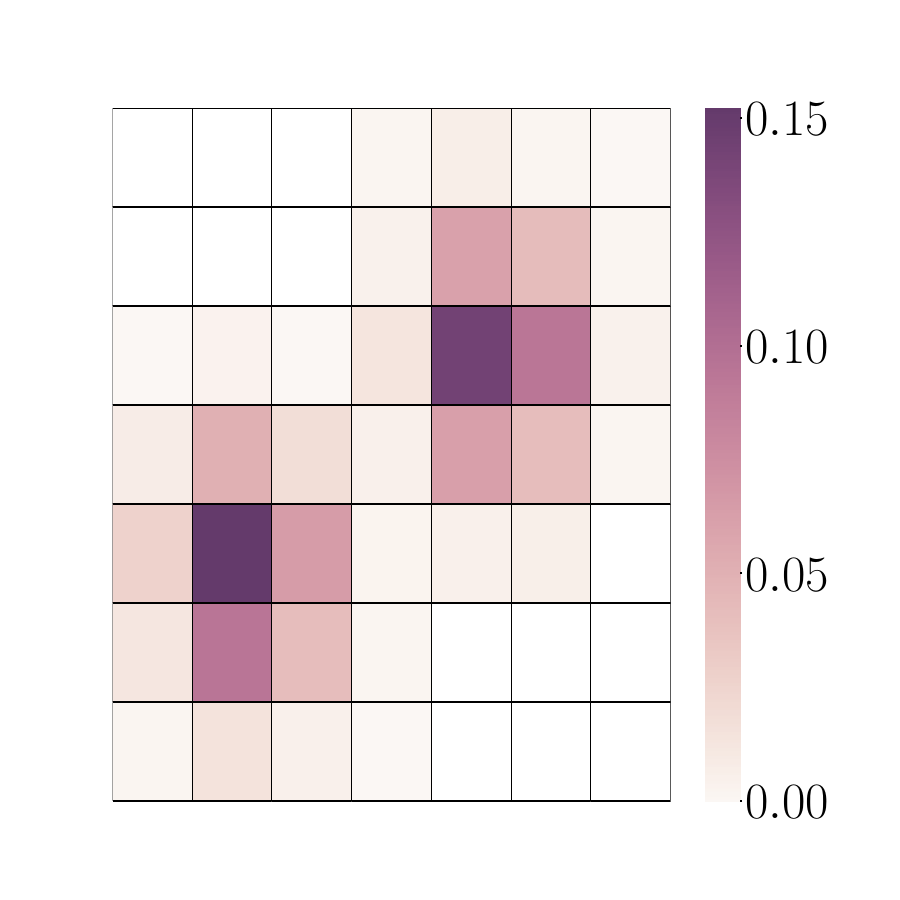} &
       \includegraphics[scale=0.25]{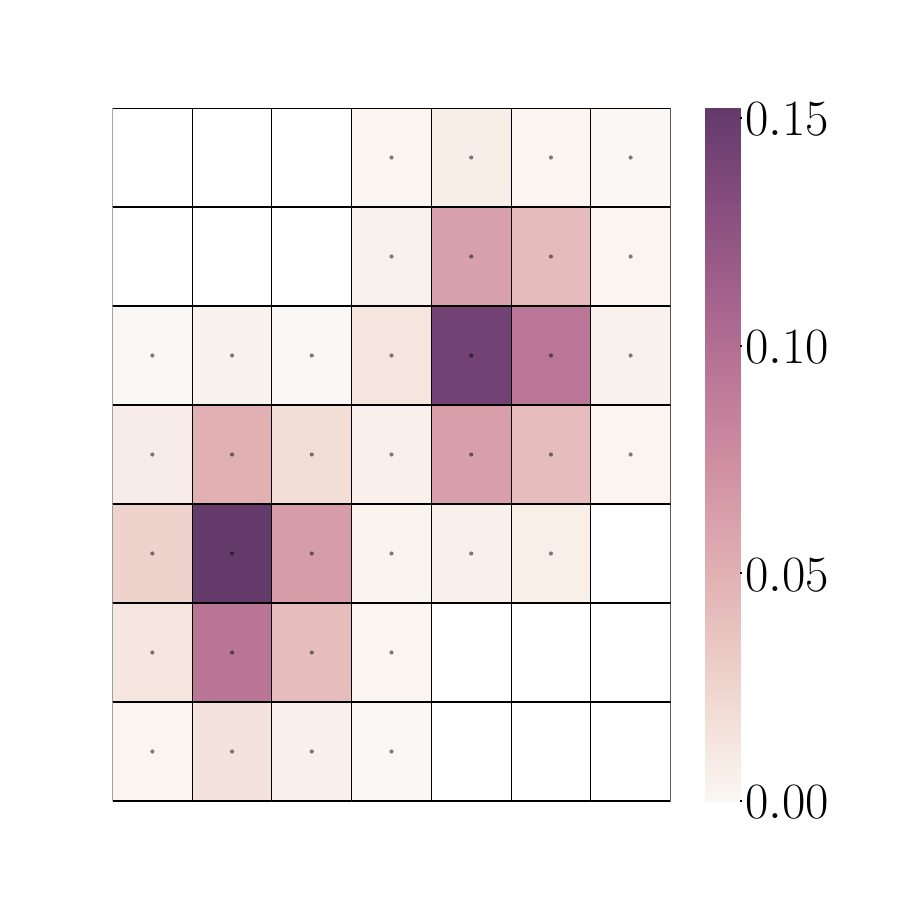} &
       \includegraphics[scale=0.25]{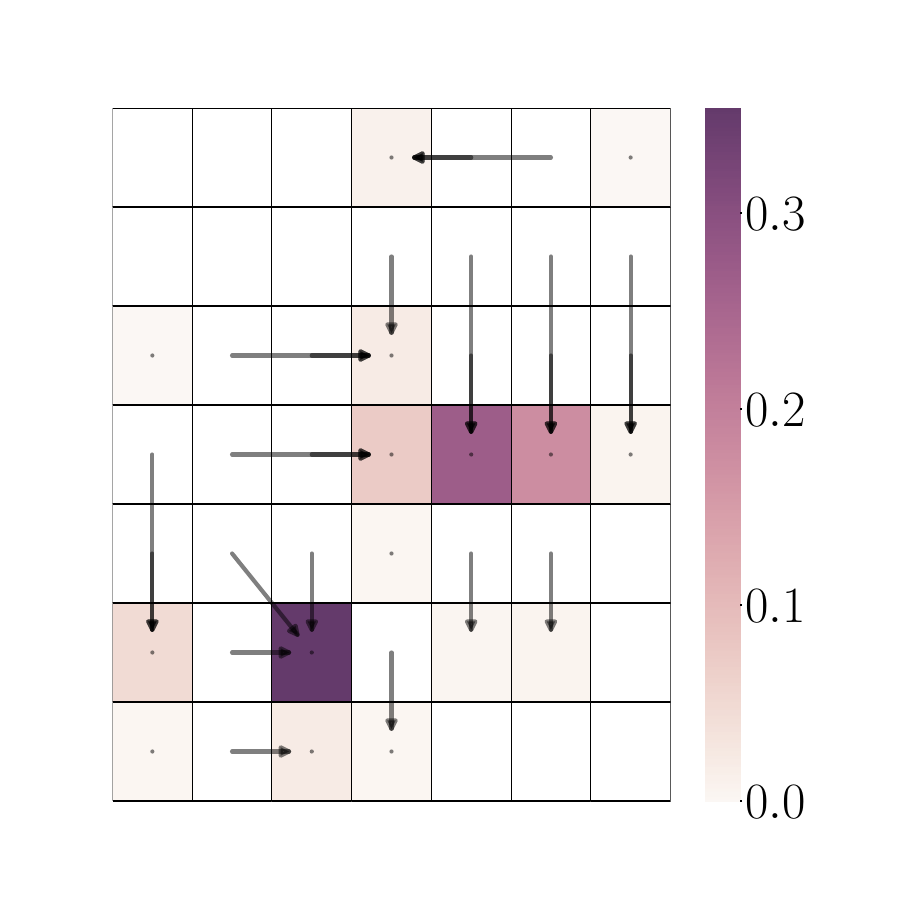} &
       \includegraphics[scale=0.25]{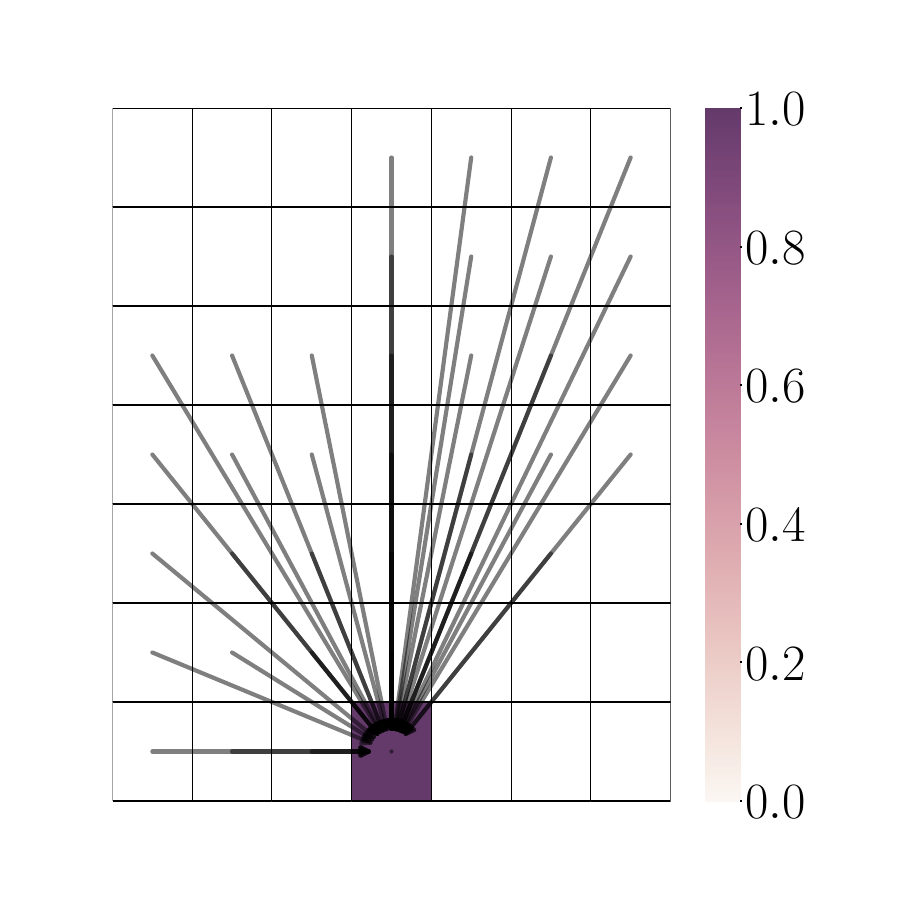}\\
       a) $P(\xb)$ & c) $P(\xb \given \pi), \alpha = 1.1$ & e) $P(\xb \given \pi), \alpha = 0.5$ & g) $P(\xb \given \pi), \alpha = 0.1$ \\
       \includegraphics[scale=0.25]{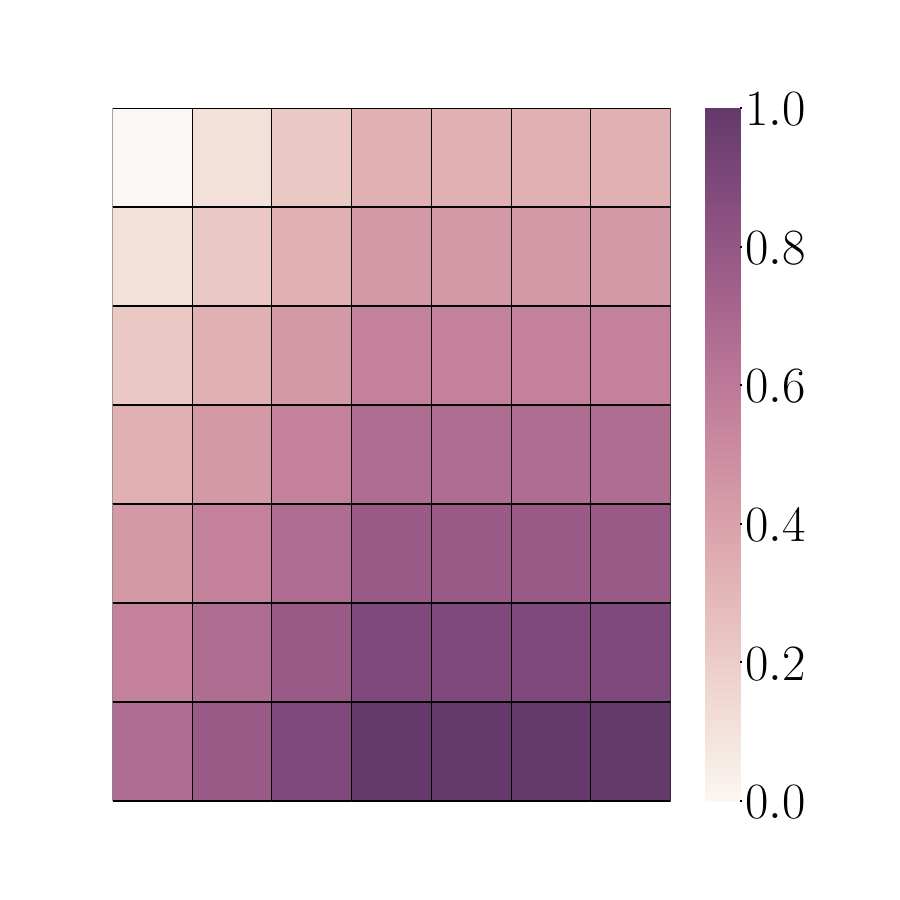}&
       \includegraphics[scale=0.25]{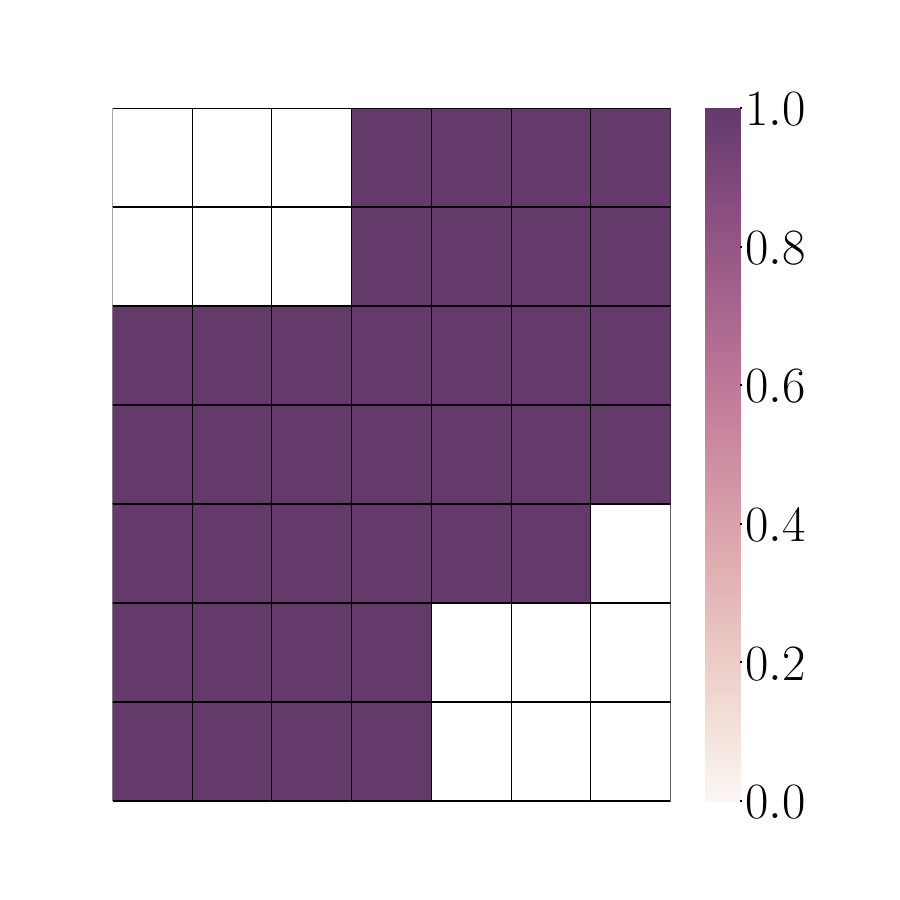}&
       \includegraphics[scale=0.25]{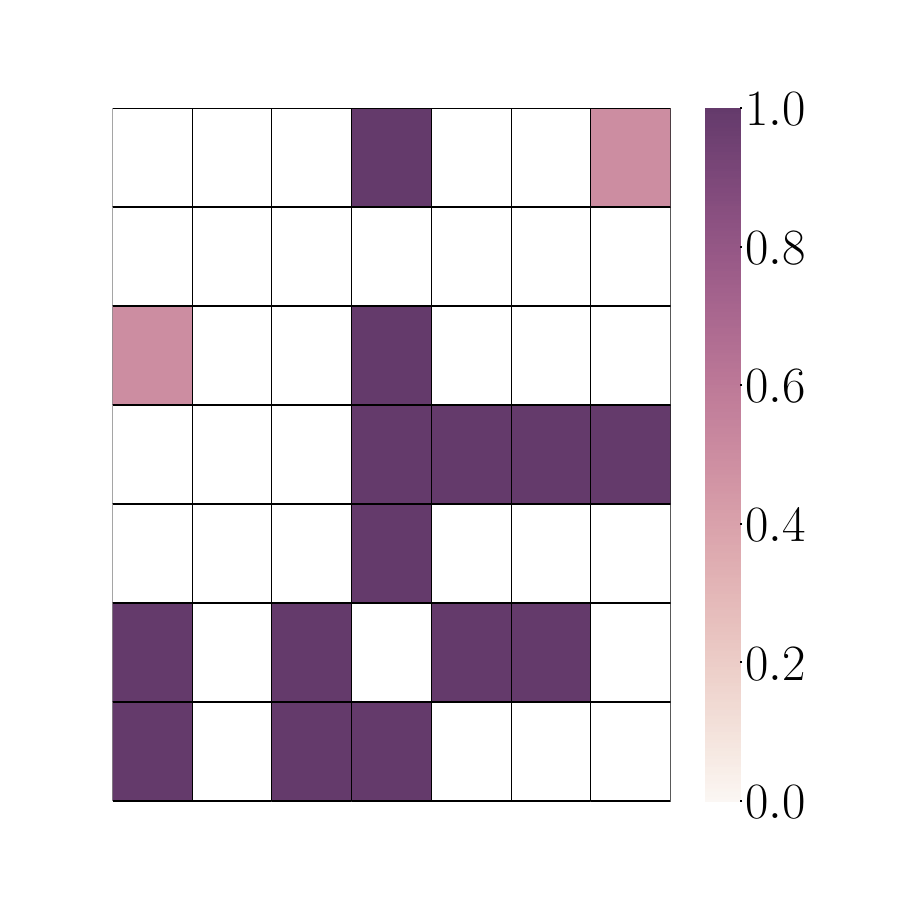}&
       \includegraphics[scale=0.25]{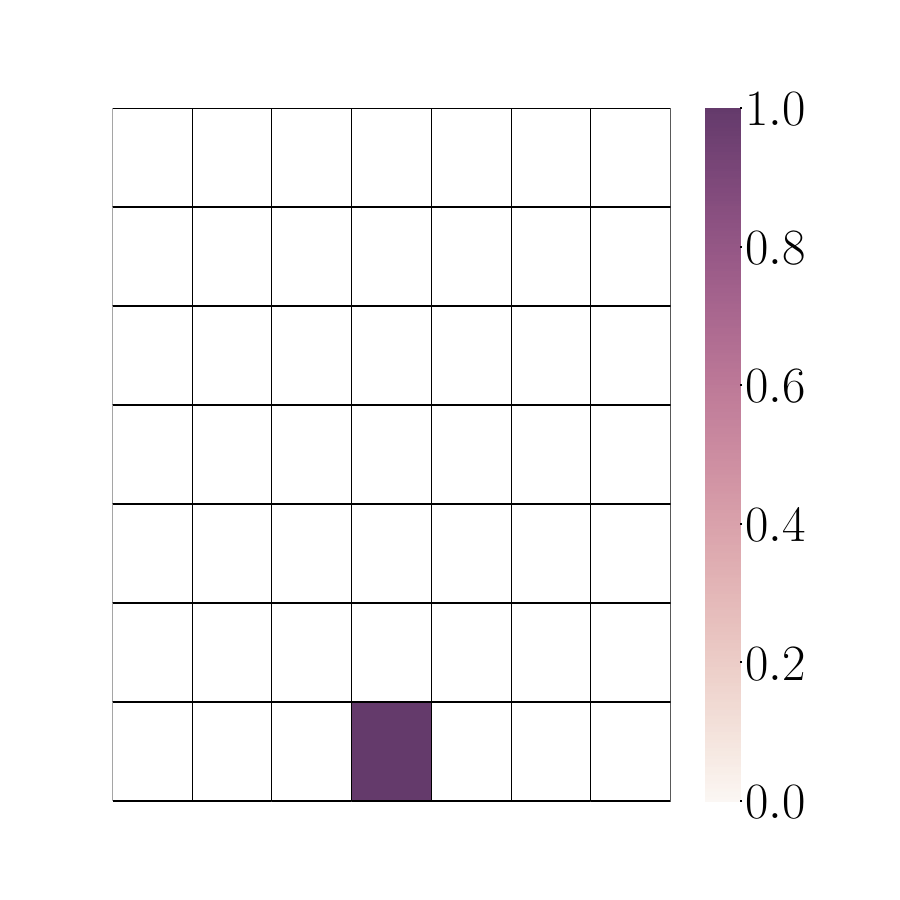}\\
       b) $P(y = 1 \given \xb)$ & d) $\pi(\xb), \alpha = 1.1$ & f) $\pi(\xb), \alpha = 0.5$ & h) $\pi(\xb), \alpha = 0.1$ 
     \end{tabu}
     \vspace{-2mm}
     \caption{Optimal decision policies and induced feature distributions.
     Panels (a) and (b) visualize $P(\xb)$ and $P(y = 1 \given \xb)$, respectively. In Panels (c-h), we set the cost to change feature values
     $c(\xb_i, \xb_j) = \alpha[|x_{i0}-x_{j0}|+|x_{i1}-x_{j1}|]$, where $\alpha$ is a given parameter and $\gamma = 0.2$.
     In all panels, each cell corresponds to a different feature value $\xb_i$ and darker colors correspond to higher values.
     As the cost of moving to further feature values for individuals decreases, the decision policy only provides positive decisions for a few $\xb$ 
     values with high $(P(y = 1 \given \xb)-\gamma)$, encouraging individuals to move to those values.
     }
     \label{fig:2dexample}
     \vspace{-3mm}
\end{figure*}
Similarly as in most previous work in strategic classification~\cite{bruckner2011stackelberg, dalvi2004adversarial, dong2018strategic, hardt2016strategic, hu2019disparate},
we consider a Stackelberg game in which the decision maker moves first before individuals best-respond. Moreover, we assume eve\-ry individual is rational 
and aims to maximize her individual benefit.
However, in contrast with previous work, we assume the decision maker shares her decision policy rather than the predictive model. 
Then, our goal is to find the (optimal) policy that maximizes utility, as defined in Eq.~\ref{eq:utility}, \ie,
\begin{equation} \label{eq:utility-maximization-1}
\pi^{*} = \argmax_{\pi} u(\pi, \gamma),
\end{equation}
under the assumption that each individual best responds.
For each individual, her best response is to change from her initial set of features $\xb_i$ to a set of features 
\begin{equation} \label{eq:benefit-maximization}
\xb_j = \argmax_{k \in [m]} b(\xb_k) - c(\xb_i, \xb_k),
\end{equation}
where $c(\xb_i, \xb_k)$ is the cost\footnote{\scriptsize In practice, the cost $c(\xb_i, \xb_k)$ for each pair of feature values may be given by a parameterized 
function.} she pays for changing from $\xb_i$ to $\xb_k$. 
%
Throughout the paper, we will assume that 
(i) it holds that $c(\xb_i, \xb_j) > 0$ for all $i \neq j$ such that $P(y \given \xb_j) \geq P(y \given \xb_i)$ and (ii) if there are ties in Eq.~\ref{eq:benefit-maximization}, the individual chooses to move to the set of features $\xb_j$ with highest $P(y \given \xb_j)$.

At a population level, this best response results into a transportation of mass between the original distribution and the induced distribution, \ie,
from $P(\xb_i)$ to $P(\xb_j \given \pi)$, as exemplified by Figure~\ref{fig:2dexample}. In particular, we can readily derive an analytical expression for 
the induced feature distribution in terms of the original feature distribution:
\begin{equation} \label{eq:induced-distribution}
P(\xb_j | \pi) = \sum_{i \in [m]} P(\xb_i) \II(\xb_j=\argmax_{k \in [m]} b(\xb_k) - c(\xb_i, \xb_k) ).
\end{equation}
%
%
Note that the transportation of mass between the original and the induced feature distribution has a natural interpretation
in terms of optimal transport theory~\cite{villani2008optimal}. More specifically, the induced feature distribution is given by
$P(\xb_j | \pi) = \sum_{i \in [m]} f_{ij}$, where $f_{ij}$ denotes the flow between $P(\xb_i)$ and $P(\xb_j | \pi)$ and it is the solution
to the following optimal transport problem:
\begin{equation*} 
\begin{split}
\underset{\{f_{ij}\}}{\text{maximize}} & \quad \sum_{i, j \in [m]} f_{ij} [b(\xb_j) - c(\xb_i, \xb_j)] \\
\text{subject to} & \quad f_{ij} \geq 0 \quad \forall i, j \quad \text{and} \quad \sum_{j \in [m]} f_{ij} = P(\xb_i).
\end{split}
\end{equation*}
%
%
%
Finally, we can combine Eqs.~\ref{eq:utility-maximization-1}-\ref{eq:induced-distribution} and rewrite our goal as follows:
\begin{align} \label{eq:utility-maximization-2}
\pi^{*} &= \argmax_{\pi} \sum_{i, j \in [m]} (P(y=1\given\xb_j)-\gamma)\pi(\xb_j) \nonumber \\
&\,\, \times \left[P(\xb_i) \II(\xb_j=\argmax_{k \in [m]} b(\xb_k) - c(\xb_i, \xb_k) ) \right]
\end{align}
where note that, by definition, $0 \leq \pi(\xb_j) \leq 1$ for all $j$, the optimal policy $\pi^{*}$ may not be unique
and, in practice, the distribution $P(\xb)$ and the conditional distribution $P(y \given \xb)$ may be
approximated using models trained on historical data (see remarks on gaming in Section~\ref{sec:preliminaries}).

Unfortunately, the following Theorem tells us that, in general, we cannot expect to find the optimal policy that maximizes utility in polynomial time 
(proven in Appendix~\ref{app:hardness} using a novel reduction of the Boolean satisfiability (SAT) problem~\cite{karp1972reducibility}):
\begin{theorem} \label{thm:hardness}
The problem of finding the optimal decision policy $\pi^{*}$ that maximizes utility in a strategic setting is NP-hard.
\end{theorem}

The above result readily implies that, in contrast with the non strategic setting, optimal decision policies are not always deterministic threshold 
rules~\cite{corbett2017algorithmic, valera2018enhancing}, \ie,
\begin{equation} \label{eq:detthresh}
    \pi^*(d = 1 \given \xb) =
     \begin{cases}
       1 &\quad \text{ if } P(y=1 \given \xb) \geq \gamma \\
       0 &\quad \text{ otherwise.}
     \end{cases}
\end{equation}
Even more, in a strategic setting, there are many instances in which the optimal decision policies are not deterministic, even under outcome monotonic costs.
For example, assume $\xb \in \{1, 2, 3\}$ with $\gamma = 0.1$,
\begin{align*}
P(\xb) &= 0.1\, \II(\xb=1) + 0.4\, \II(\xb=2) + 0.5\, \II(\xb=3) \\
P(y = 1 \given \xb) &= 1.0\, \II(\xb=1) + 0.7\, \II(\xb=2) + 0.4\, \II(\xb=3) \\
%
%
  c(\xb_i, \xb_j) &= \begin{bmatrix} 0.0 & 0.0 & 0.0 \\
                                  0.3 & 0.0 & 0.0 \\
                                  1.2 & 0.3 & 0.0
                      \end{bmatrix}
\end{align*}
In the non-strategic setting, the optimal policy is clearly $\pi^*(d = 1 \given \xb = 1) = 1$, $\pi^*(d = 1 \given \xb = 2) = 1$ and
$\pi^*(d = 1 \given \xb = 3) = 1$.
%
%
However, in the strategic setting, a brute force search reveals that the optimal policy is stochastic and it is given by $\pi^*(d = 1 \given \xb = 1) = 1$,
$\pi^*(d = 1 \given \xb = 2) = 0.7$ and $\pi^*(d = 1 \given \xb = 3) = 0$, inducing a transportation of mass from $P(\xb = 3)$ to $P(\xb = 2 \given \pi)$ and from $P(\xb = 2)$ to $P(\xb = 1 \given \pi)$.
Moreover, note that the optimal policy in the strategic setting achieves higher utility than its counterpart in the non-strategic setting.

%
%

\vspace{-2mm}
\section{Outcome Monotonic Costs}
\label{sec:monotonic}
\vspace{-1mm}
In this section, we show that, if the cost in\-di\-vi\-duals pay to change features satisfies \emph{outcome monotonicity}~\cite{milli2018social, hu2019disparate}, 
%
%
we can narrow down the search for the optimal policy to a particular family of decision policies with a set of desirable properties.
%
%
%
A cost satisfies outcome mo\-no\-to\-ni\-ci\-ty if improving an individual'{}s outcome requires increasing amount of effort, \ie, 
%
%
\begin{align*}
P(y=1 \given \xb_i) &< P(y=1 \given \xb_j) < P(y=1 \given \xb_k) \\
& \Leftrightarrow \left[c(\xb_i, \xb_j) < c(\xb_i, \xb_k) \right] \wedge \left[ c(\xb_j, \xb_k) < c(\xb_i, \xb_k) \right]
\end{align*} 
and worsening an individual'{}s outcome requires no effort, \ie, $P(y = 1 \given \xb_i) > P(y = 1 \given \xb_j) \Leftrightarrow c(\xb_i, \xb_j) = 0$.
Here, without loss of generality, we will index the feature values in decreasing order with respect to their corresponding outcome, \ie, $i < j \Rightarrow P(y \given \xb_i) \geq P(y \given \xb_j)$.

Given any instance of the utility maximization problem, as defined in Eq.~\ref{eq:utility-maximization-1}, it is easy to show that the optimal policy will 
always decide positively about the feature value with the highest outcome\footnote{\scriptsize As long as $P(y \given \xb_1) > \gamma$.}, \ie, $\pi^{*}(\xb_1) = 1$,
and negatively about the feature values with outcome lower than $\gamma$, \ie, $P(y \given \xb_i) < \gamma \Rightarrow \pi^{*}(\xb_i) = 0$.
However, if the cost in\-di\-vi\-duals pay to change features satisfies outcome monotonicity, we can further characterize a
particular family of decision policies that is guaranteed to contain a policy that achieves the optimal utility. 
In particular, we start by showing that there exists an optimal policy that is outcome monotonic (proven in 
Appendix~\ref{app:outcome-monotonicity}):
\begin{proposition} \label{prop:outcome-monotonicity}
Let $\pi^{*}$ be an optimal policy that maximizes utility. If the cost $c(\xb_i, \xb_j)$ is outcome monotonic then there 
exists an outcome monotonic policy $\pi$ such that $u(\pi, \gamma) = u(\pi^{*}, \gamma)$.
\end{proposition}

In the above, note that, given an individual with an initial set of features $\xb_i$, an outcome monotonic policy always induces a 
best response $\xb_j$ such that $P(y \given \xb_j) \geq P(y \given \xb_i)$. Otherwise, a contradiction would occur since, by assumption, 
it would hold that $\pi(\xb_i) \geq \pi(\xb_j)$ and $\pi(\xb_j) \geq \pi(\xb_j) - c(\xb_i,\xb_j)$.
%
%
%
%
%
Next, we consider additive costs, \ie, $c(\xb_i, \xb_j) + c(\xb_j, \xb_k) = c(\xb_i, \xb_k)$,
and afterwards move on to subadditive costs, \ie, $c(\xb_i, \xb_j) + c(\xb_j, \xb_k) \geq c(\xb_i, \xb_k)$

%
%
%

\xhdr{Additive costs} If the cost is additive, we first show that we can narrow down the search for the optimal policy to the policies $\pi$ that 
satisfy that
\begin{equation} \label{eq:outcome-monotonic-binary-policy}
\pi(\xb_i)=\pi(\xb_{i-1}) \vee \pi(\xb_i)=\max(0, \pi(\xb_{i-1})-c(\xb_i,\xb_{i-1}))
\end{equation}
for all $i > 1$ such that $P(y\given \xb_i)>\gamma$. In the remainder, we refer to any policy with this property as an outcome monotonic binary policy. More formally, 
we have the fo\-llowing Theorem (proven in Appendix~\ref{app:outcome-monotonic-binary-policy}):
\begin{theorem} \label{thm:outcome-monotonic-binary-policy}
Let $\pi^{*}$ be an optimal policy that maximizes utility. If the cost $c(\xb_i, \xb_j)$ is additive and outcome monotonic then there exists an 
outcome monotonic binary policy $\pi$ such that $u(\pi, \gamma) = u(\pi^{*}, \gamma)$.
\end{theorem}
Moreover, we can further characterize the best-responses of individuals under outcome monotonic 
binary policies and additive costs (proven in Appendix~\ref{app:best-response}):
\begin{proposition} \label{prop:best-response}
Let $\pi$ be an outcome monotonic binary policy, $c(\xb_i, \xb_j)$ be an additive and outcome monotonic cost,
$\xb_i$ be an individual'{}s initial set of features, and define $j = \max \{k \given k \leq i, \pi(\xb_k) = 1 \vee \pi(\xb_k) = \pi(\xb_{k-1})\}$.
If $P(y\given\xb_i)>\gamma$, the individual'{}s best response is $\xb_j$ and, if $P(y\given\xb_i)\leq\gamma$, the individual'{}s 
best response is $\xb_j$ if $\pi(\xb_j)\geq c(\xb_i,\xb_j)$ and $\xb_i$ otherwise.
\end{proposition}
%
%
%
%
This proposition readily implies that $P(\xb_i \given \pi) = 0$ for all $\xb_i$ such that $\pi(\xb_i) \neq \pi(\xb_{i-1})$ with $\pi(\xb_i)>0$. Therefore, it lets us think of the feature values $\xb_i$ with $\pi(\xb_i) = \pi(\xb_{i-1})$ as \emph{blocking} states and those with 
$\pi(\xb_i) \neq \pi(\xb_{i-1})$ as \emph{non-blocking} states.
Moreover, the above results facilitate the development of a highly effective heuristic search algorithm based on dynamic
programming to find close to optimal (outcome monotonic binary) policies in polynomial time, which we describe later in this section.
%
%
%
%
%

\xhdr{Subadditive costs}
If the cost is subadditive, we can show that we need to instead narrow down the search for the optimal policy to the policies $\pi$ that 
satisfy that
\begin{equation} \label{eq:multiple-steps-policy}
\pi(\xb_i) = \pi(\xb_{i-1}) \vee \bigvee_j \pi(\xb_i)=\max(0, \pi(\xb_{i-1})-c(\xb_j, \xb_{i-1}))
\end{equation}
for all $i > 1$ such that $P(y\given \xb_i)>\gamma$ and $j = i, \ldots k$ with $k = \max\{ j \given \pi(\xb_{i-1}) - c(\xb_j, \xb_{i-1}) > 0 \}$. 
More~for\-mally, we have the following proposition, which can be easily shown using a similar reasoning as the one used in the proof of 
Theorem~\ref{thm:outcome-monotonic-binary-policy}:
\begin{proposition} \label{prop:outcome-monotonic-binary-policy}
Let $\pi^{*}$ be an optimal policy that maximizes utility. If the cost $c(\xb_i, \xb_j)$ is subadditive and outcome monotonic then there exists an outcome 
monotonic policy $\pi$ satisfying Eq.~\ref{eq:multiple-steps-policy} such that $u(\pi, \gamma) = u(\pi^{*}, \gamma)$.
\end{proposition}
Similarly as in the case of additive costs, it is possible to characterize the best response of the individuals\footnote{\scriptsize In this
case, however, each possible decision policy value blocks zero, one or more feature values.} and adapt the above mentioned heuristic 
search algorithm to find optimal (outcome monotonic binary) policies with subadditive costs, however, the resulting algorithm is rather
impractical due to its complexity and therefore we omit the details.

\begin{algorithm}[t!]
\small
\caption{\textsc{DynamicProgramming}: It searches for the optimal decision policy that maximizes utility under additive and outcome monotonic costs.}
\label{alg:dp}
  \begin{algorithmic}[1]
    \REQUIRE Number of feature values $m$, constant $c$, distributions $\Pb = [P(\xb_i)]$ and $\Qb = [P(y \given \xb_i)]$, and cost $\Cb=[c(\xb_i, \xb_j)]$ 
      \vspace{1mm}
      \STATE $\Pi \gets \textsc{InitializeSubpolicies}{()}$
      \vspace{1mm}
      
      \STATE $s \gets 1$
      \REPEAT
          \STATE $r, \Pi, F \gets \textsc{ComputeBaseSubpolicies}{(\Cb, \Pb, \Qb, \pi_{2, 1}(\xb_s))}$
       \FOR{$i = r-1, \ldots, s+1$}
            \FOR{$j = i-1, \ldots, s$}
    \IF{$c(\xb_{i-1},\xb_j) > \pi_{2,1}(\xb_s)$}{\STATE $\mathbf{continue}$} \ENDIF
          \STATE $V(i, j) = m$
          \STATE $\sigma \gets c(\xb_{i-1},\xb_j)$
          \STATE $G \gets (P(y \given \xb_j)-c) \sum_{j \,:\, j \leq k < i} P(\xb_k)$
        \STATE $\pi', F', v' \gets \textsc{Lower}{(\pi_{i+1,i}, F(i+1, i), \sigma)} + G)$
        \vspace{1mm}
          \IF{$F(i+1, j) \geq F'$ \AND $c(\xb_i,\xb_j) \leq \pi_{2,1}(\xb_s)$}
          \STATE $F(i, j) \gets F(i+1, j)$
          \STATE $\pi_{i,j}(\xb_i) \gets \pi_{i,j}(\xb_{i-1}) - c(\xb_i, \xb_{i-1})$
          \FOR{$l=i+1, \cdots, m$}
            \STATE $\pi_{i, j}(\xb_l) = \pi_{i+1, j}(\xb_l)$
          \ENDFOR
      \STATE $V(i, j) = V(i+1, j)$
        \ELSE
          \STATE $F(i, j) \gets F'$
          \STATE $\pi_{i,j}(\xb_i) \gets \pi_{i,j}(\xb_{i-1}) $
          \FOR{$l=i+1, \cdots, m$}
            \STATE $\pi_{i, j}(\xb_l) = \pi'(\xb_l)$
          \ENDFOR
          \STATE $V(i, j) \gets \min(v', V(i+1, i))$
        \ENDIF
    \vspace{1mm}
            \ENDFOR
          \ENDFOR
    \vspace{1mm}
    \FOR{$l = s, \ldots, V(s+1, s)$}
      \STATE $\pi(\xb_l) \gets \pi_{s+1, s}(\xb_l)$
    \ENDFOR
          \STATE $s \gets V(s+1, s)$
      \UNTIL{$V(s+1,s) = m$}
      \STATE \mbox{\bf Return} $\pi, u(\pi, \gamma)$
  \end{algorithmic}
\end{algorithm}
\begin{figure*}[t]
     \centering
     \subfloat[Optimal policy $\pi^{*}(\xb)$]{
         \centering
         \includegraphics[scale=0.19]{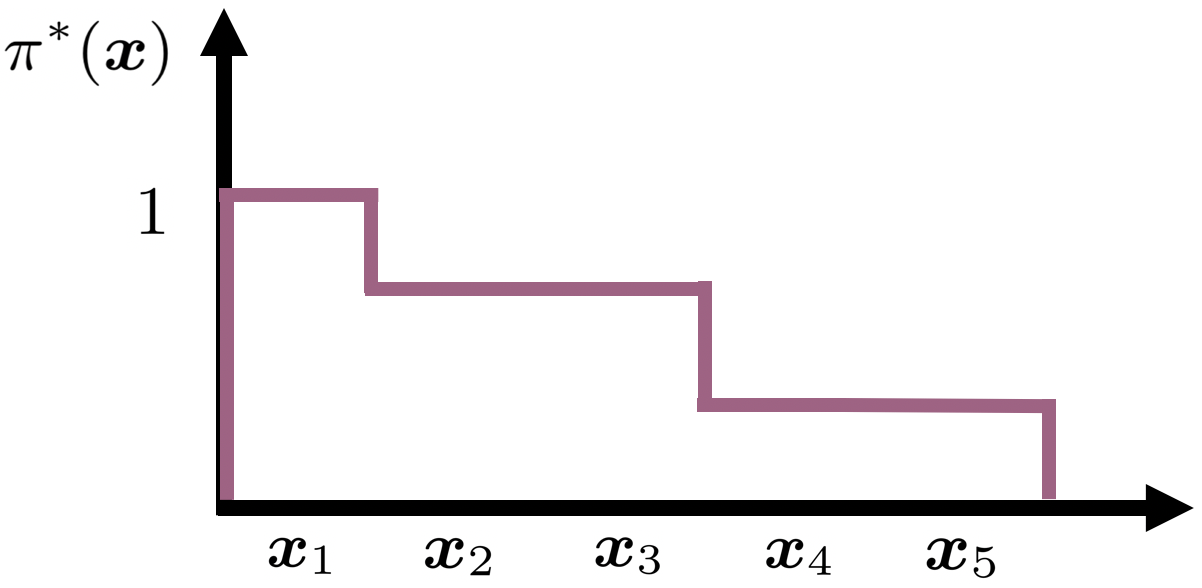}
     } \hspace{-5mm}
     \subfloat[Subpolicy $\pi_{5,3}(\xb)$]{
         \centering
         \includegraphics[scale=0.19]{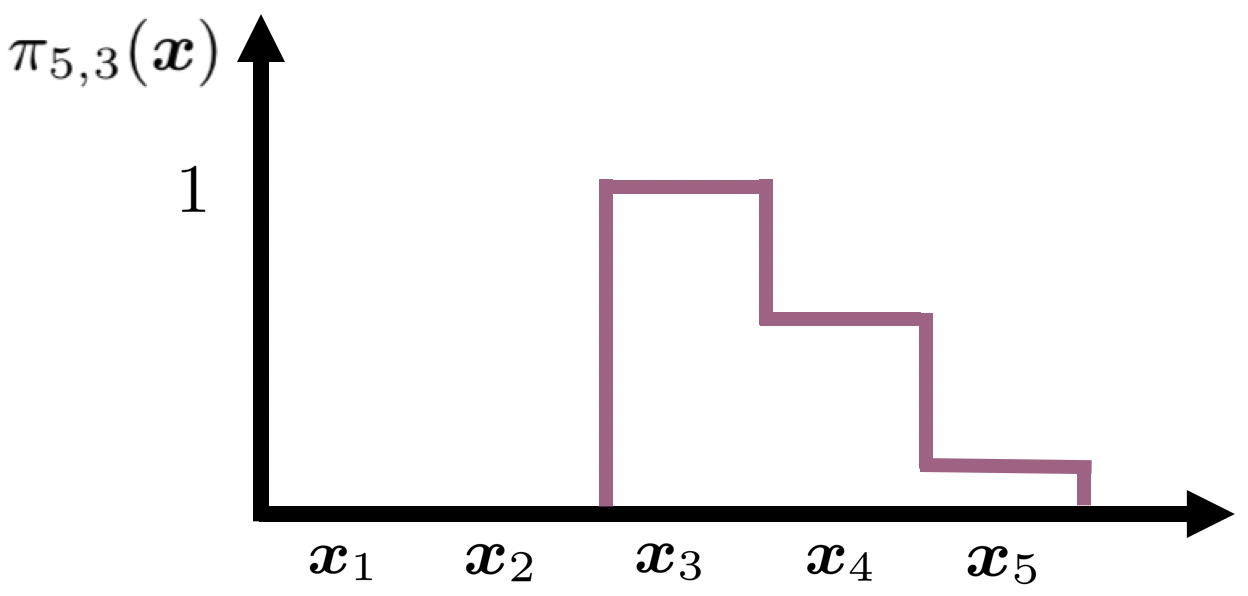}
     } \hspace{-5mm}
     \subfloat[Subpolicy $\pi_{4,2}(\xb)$]{
         \centering
         \includegraphics[scale=0.19]{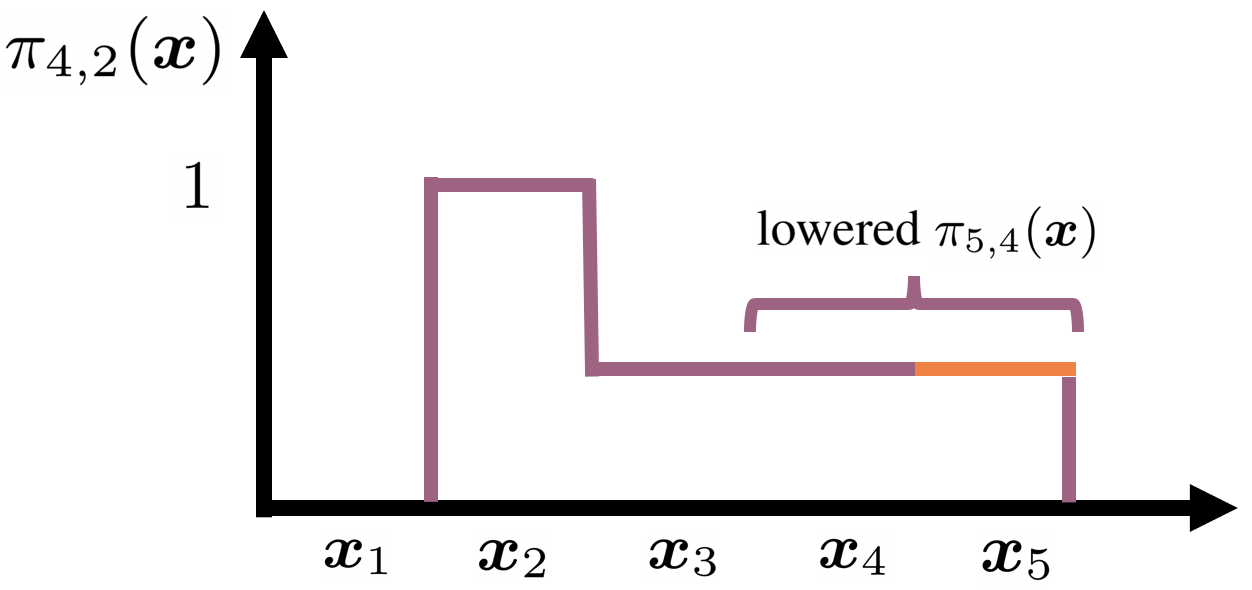}
     } \hspace{-5mm}
     \subfloat[Subpolicy $\pi_{2, 1}(\xb)$]{
         \centering
         \includegraphics[scale=0.19]{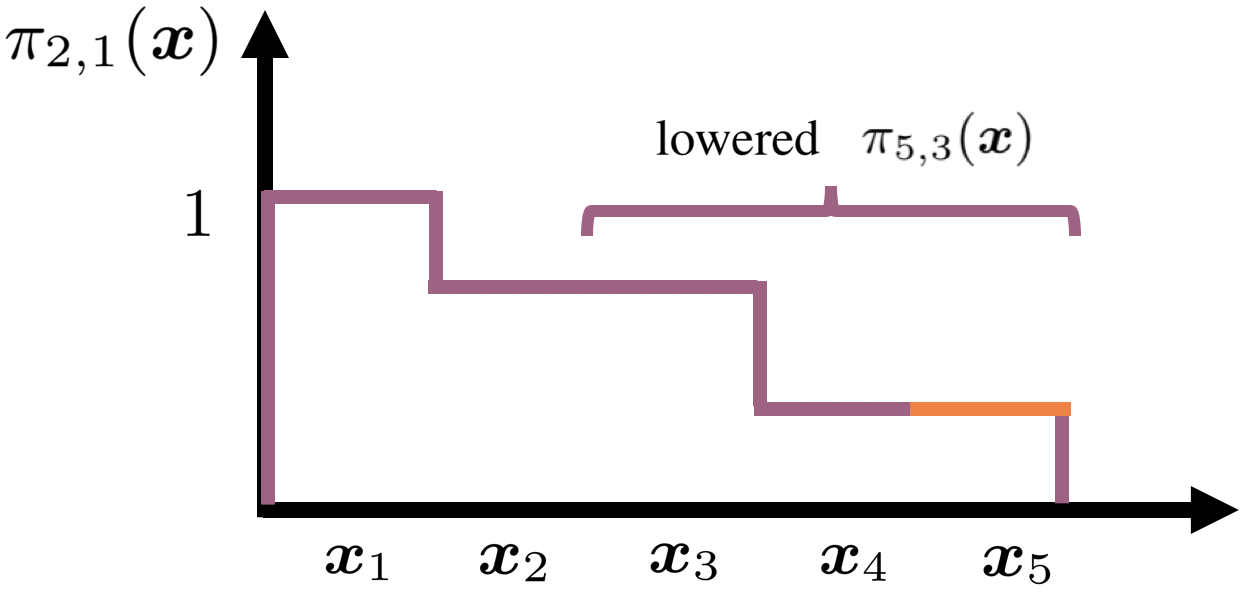}
     } 
     \caption{Optimal policy and subpolicies after Algorithm~\ref{alg:dp} performs its first round. Panel (a) shows the optimal subpolicy $\pi^{*}(\xb)$, which contains
     blocking states in $\xb_3$ and $\xb_5$. Panel (b) shows the subpolicy $\pi_{5,3}(\xb)$, which is a base subpolicy that can be computed without recursion. 
     Panel (c) shows the subpolicy $\pi_{4,2}(\xb)$, which contains a blocking state in $\xb_4$ and uses a lowered version of the subpolicy $\pi_{5,4}(\xb)$ to set the 
     feature value $\xb_5$.
     Since $\pi_{4,2}(\xb_4)-c(\xb_5,\xb_4)<0$, this value is set equal to $\pi_{4,2}(\xb_4)$.
     Panel (d) shows the subpolicy $\pi_{2,1}(\xb)$, which contains a blocking state in $\xb_3$ and uses a lowered version of the subpolicy $\pi_{5,3}(\xb)$
     to set the feature values $\xb_4$ and $\xb_5$. 
     Since in $\pi_{2,1}(\xb)$, the feature value $\xb_5$ became negative and was set as blocking, the algorithm will perform a second round, starting from $\xb_3$, which is the last blocking 
     state before $d=4$.}
     \label{fig:subpolicies}
\end{figure*}

\xhdr{Dynamic programming algorithm}
%
%
The key idea behind our algorithm (refer to Algorithm~\ref{alg:dp}) is to recursively create a set of decision subpolicies $\Pi = (\pi_{i, j}(\xb_k))$ where $i,j=1, \ldots, m$ with $j < i$, $k=j, \ldots, m$, 
which we later use to build the entire decision policy $\pi$. 
Moreover, depending on the structure of the costs and feature and label distributions, we may need to perform several rounds and, in each round, create a new set of decision 
subpolicies, which are used to set only some values of the decision policy (lines 31--33). 

More specifically, in each round, we proceed in de\-crea\-sing order of $i$ and $j$ (lines 5--6) until the feature value index $s$, which is computed in the previous round (lines 20, 27). 
For each subpolicy $\pi_{i, j}$: 
(i) we fix $\pi_{i,j}(\xb_j) = \pi_{2,1}(\xb_s)$, $\pi_{i,j}(\xb_k) = \pi(\xb_{k-1}) - c(\xb_k, \xb_{k-1})$ for all $j < k < i$ and $\pi_{i,j}(\xb_k) = 0$ for all $k$ such that $P(y \given \xb_k) \leq \gamma)$ 
(line 4); 
(ii) we decide whether to \emph{block} or not to block the feature value $\xb_i$, \ie, set $\pi_{i,j}(\xb_i)$ to either $\pi_{i,j}(\xb_{i-1})$ or $\pi_{i,j}(\xb_{i-1}) - c(\xb_{i}, \xb_{i-1})$, based on previously computed subpolicies within the round (lines 13--14); and, (iii) if we decide to block the feature value $\xb_i$, we set the remaining policy values by \emph{appending}
the \emph{best} of these previously computed subpolicies in terms of overall utility (lines 15--19 and 22--26). 
Here, note that there is a set of base subpolicies, those with $i = r$ where $r=\max \{k: P(y \given \xb_k) > \gamma\}$ and $1 - c(\xb_{i-1},\xb_j) \geq 0$, which can be computed directly, without 
recursion (line 4).
Moreover, if we decide to block $\xb_i$ in a subpolicy $\pi_{i,j}$, we need to lower the values of the previously computed subpolicies down (line 13) by $\sigma = c(\xb_{i-1}, \xb_j)$ before appending them so that $\pi_{i,j}(\xb_i)=\pi_{2,1}(\xb_s) - c(\xb_{i-1}, \xb_j)$ eventually. 
However, some of these values may become negative and are thus decided to be blocking sates, \ie, $\pi'(\xb_k)=\pi_{i+1, i}(\xb_d)-\sigma\ \forall k: r\geq k>d$ where $d=\max \{l: \pi_{i+1, i}(\xb_l)-\sigma\geq 0\}$. If during this procedure the lowered policy makes some individual change their best-response, the 
policy values starting from the last blocking state $v'$ before $d$ will be revisited in another round (line 35).
Figure~\ref{fig:subpolicies} shows several examples of subpolicies $\pi_{i,j}$, showing the lowering procedure.
%
%

Within the algorithm, the function \textsc{InitializeSubpolicies}$()$ initializes the subpolicies $\Pi = \{\pi_{i,j}\}$,
\textsc{ComputeBaseSubpolicies}$(\ldots)$ computes $r=\max \{k: P(y \given \xb_k) > \gamma\}$, the base subpolicies and their utilities,
%
%
and
\textsc{Lower}$(\pi_{i+1,i}, F(i+1, i), \sigma)$ computes a policy $\pi'$ with $\pi'(\xb_k) = \pi_{i+1, i}(\xb_k) - \sigma$ if that quantity is non-negative and $\pi'(\xb_k)=\pi_{i+1, i}(\xb_d)-\sigma$ otherwise,
its corresponding utility $F'$, 
and calculates the index $v'$ of the last blocking state before $d$ as described in the previous paragraph.

As mentioned above, the algorithm might need more than one round to terminate.
Since each round consists of one dynamic programming execution, an array of utility values of all subpolicies needs to be computed, having a size of $\Ocal(m^2)$, considering that each state variable $i,j$ takes values from the set $\{1,2,..,m\}$.
Given an outcome monotonic binary policy $\pi$, according to Proposition~\ref{prop:best-response}, we can easily characterize the best-response of each individual and it can be easily seen that the overall utility $u(\pi,\gamma)$ can be computed with a single pass over the feature values.
Therefore, computing each entry's value in the aforementioned array takes $\Ocal(m)$ time, leading to a total round complexity of $\Ocal(m^3)$.

Now, consider the total number of rounds. It can be observed that a second round is executed iff $s\neq m$ at the end of the first one, implying that at least one feature value was blocked since the value of $V(i,j)$ might get altered only when choosing to block a feature value because of the \textsc{Lower} operation.
Therefore, we can deduce that during each round ending with $s\neq m$, at least one feature value gets blocked, leading to a $\Ocal(m)$ bound on the total number of rounds.
As a consequence, the overall complexity of the algorithm is $\Ocal(m^4)$.

\vspace{-2mm}
\section{General costs}
\label{sec:general}
\vspace{-1mm}
In this section, we first show that, under no assumptions on the cost people pay to change features, the optimal policy may violate outcome 
monotonicity. 
Then, we introduce an efficient iterative algorithm that it is guaranteed to terminate and find locally optimal decision policies.
Finally, we propose a variation of the algorithm that can significantly reduce its running time when working with real data.

\xhdr{There may not exist an optimal policy that satisfies outcome monotonicity under general costs}
Our starting point is the toy example introduced at the end of Section~\ref{sec:formulation}. Here, we just mo\-di\-fy the cost individuals
pay to change features so that it violates outcome monotonicity of the costs. More specifically, assume $\xb\in\{1,2,3\}$ with $\gamma=0.1$,
\begin{align*}
P(\xb) &= 0.1\, \II(\xb=1) + 0.4\, \II(\xb=2) + 0.5\, \II(\xb=3) \\
P(y = 1 \given \xb) &= 1.0\, \II(\xb=1) + 0.7\, \II(\xb=2) + 0.4\, \II(\xb=3) \\
%
%
  c(\xb_i, \xb_j) &= \begin{bmatrix} 0.0 & 0.2 & 0.3 \\
                                  0.3 & 0.0 & 0.7 \\
                                  1.2 & 1.1 & 0.0
                      \end{bmatrix}
\end{align*}

Now, in the strategic setting, it is easy to see that every policy given by $\pi^*(d = 1 \given \xb = 1) = 1$,
$\pi^*(d = 1 \given \xb = 2) \leq 0.7$ and $\pi^*(d = 1 \given \xb = 3) = 1$ is optimal and induces a transportation of mass from $P(\xb = 2)$ to $P(\xb = 1 \given \pi)$.
Therefore, optimal policies are not necessarily outcome monotonic under general costs.
%
\begin{algorithm}[t]
\small
\caption{\textsc{Iterative}: It approximates the optimal decision policy that maximizes utility under general cost.}
\label{alg:iterative}
  \begin{algorithmic}[1]
    \REQUIRE Number of feature values $m$, constant $c$, distributions $\Pb = [P(\xb_i)]$ and $\Qb = [P(y \given \xb_i)]$, and cost $\Cb=[c(\xb_i, \xb_j)]$ 
      \STATE $\pi \gets \textsc{InitializePolicy}{()}$
      \REPEAT
      \STATE $\pi'{} \gets \pi$
      \FOR{$i = 1, \ldots, m$}
           \STATE $\pi(\xb_i) \gets \textsc{solve}(i,\pi,\Cb,\Pb,\Qb)$
      \ENDFOR
      \UNTIL{$\pi = \pi'{}$}
      \STATE \mbox{\bf Return} $\pi'{}, u(\pi'{}, c)$
  \end{algorithmic}
\end{algorithm}

\xhdr{An iterative algorithm for general costs}
The iterative algorithm is based on the following key insight:
fix the decision policy $\pi(\xb)$ for all feature values $\xb = \xb_k$ except $\xb = \xb_i$. Then, Eq.~\ref{eq:utility-maximization-2} reduces to
searching over a polynomial number of values for $\pi(\xb_i)$. 

%
Exploiting this insight, the iterative algorithm proceeds ite\-ra\-ti\-vely and, at each each iteration, it optimizes the decision policy for
each of the feature values while fixing the decision policy for all other values.
Algorithm~\ref{alg:iterative} summarizes the iterative algorithm. Within the algorithm, \textsc{InitializePolicy}$()$ initializes the decision policy
to $\pi(\xb) = 0$ for all $\xb$,
\textsc{Solve}$(i, \pi, \Cb, \Pb,\Qb)$ finds the best policy $\pi(\xb_i)$ for $\xb_i$ given $\pi(\xb_k)$ for all $\xb_k \neq \xb_i$, $\Cb = [c(\xb_i, \xb_j)]$, 
$\Pb = [P(\xb_i)]$ and $\Qb = [P(y \given \xb_i)]$ by searching over a polynomial number of values.
In practice, we proceed over feature values in decreasing order with respect to $P(y \given \xb_i)$ because we have observed it improves performance. 
However, our theoretical results do not depend on such ordering.
%
In the following, we refer to lines 2-7 of Algorithm~\ref{alg:iterative} as one \textit{iteration} and line 5 as one \textit{step}.


%
%
%

\xhdr{Theoretical guarantees of the iterative algorithm}
We start our theoretical analysis with the following Proposition, which shows that our algorithm is guaranteed to terminate and that the number of steps is, 
at most polynomial, in the number of feature values $m$ (proven in Appendix~\ref{app:termination}):
\begin{proposition} \label{prop:termination}
Algorithm~\ref{alg:iterative} terminates after at most $m^{1+{1 \over \bar{u}}} -1$ steps, where $\bar{u}$ is the greatest common denominator
of all elements in the set $A = \lbrace c(\xb_i, \xb_j) - c(\xb_i, \xb_k) \given \xb_i, \xb_j, \xb_k \in \{1, \ldots, m\} \rbrace \cup 1$\footnote{\scriptsize The common
denominator $\bar{u}$ satisfies that ${a \over \bar{u}} \in \mathbb{Z} \,\, \forall a \in A \cup \lbrace 1 \rbrace$. Such $\bar{u}$ exists if and only if ${a \over b}$ is
rational $\forall a,b \in A$.}. 
\end{proposition}
It readily follows that, at each step, our iterative algorithm is guaranteed to find a better policy $\pi$, \ie, $u(\pi, \gamma) > u(\pi'{}, \gamma)$. This is
because \textsc{Solve}$(i, \pi, \Cb, \Pb)$ always returns a better policy $\pi$ and, by definition, at the end of each step, $\pi \neq \pi'$.
As a direct consequence of the above results, we can conclude that Algorithm~\ref{alg:iterative} finds locally optimal decision policies in polynomial time.
%

%
Moreover, we can characterize the computational complexity of the algorithm as follows.
At each iteration, the algorithm calls \textsc{Solve} $m$ times and, within \textsc{Solve}, there are $\Ocal(m)$ candidate values 
for $\pi(\xb_i)$ when $\pi(\xb_k)$ is fixed for all $\xb_k \neq \xb_i$ and they can all be evaluated in $\Ocal(m^2)$.
Therefore, the iteration complexity of Algorithm~\ref{alg:iterative} is $\Ocal(m^3)$.

\xhdr{Speeding up the iterative algorithm in the presence of non-actionable features}
In this section, we discuss a highly effective strategy to speed up the iterative algorithm whenever some of the features are 
non-actionable, which is amenable to parallelization.
As an example, assume there is an \textit{Age Group} feature which takes values $\{<30, 30-60, >60\}$.
Now, consider two individuals with initial feature values $\xb_i, \xb_j$ such that $\xb_{i, Age Group}=\ <30$ and $\xb_{j, Age Group}=\ >60$.
Since individuals cannot change their age, it holds that $c(\xb_i, \xb_j) = c(\xb_j, \xb_i) = \infty$.
Let $\Gcal$ be an undirected graph where each node $v_i$ represents a feature value $\xb_i$ and there is an edge $e_{i,j}$
between two nodes $v_i$ and $v_j$ iff $c(\xb_i, \xb_j)\leq 1 \vee c(\xb_j, \xb_i)\leq 1$.
Then, if there are non-actionable features, it is easy to see that the graph $\Gcal$ may be composed of several independent 
connected components. 
Assume $v_i$ and $v_j$ belong to two different connected components. 
Then, whatever value is picked for $\pi(\xb_i)$, the best-response of individuals with initial features $\xb_j$ will never 
be $\xb_i$ since $\pi(\xb_i)\leq 1 \Rightarrow \pi(\xb_i)-c(\xb_j,\xb_i) \leq 1-c(\xb_j,\xb_i)<0 \leq \pi(\xb_j)$ and therefore 
$\xb_j$ will always be a better response.
Similarly, the best-response of individuals with initial features $\xb_i$ will never be $\xb_j$ independently of the value of $\pi(\xb_j)$.
As a consequence, we can find the values of the optimal policy by running the iterative algorithm independently on each independent component.

\vspace{-2mm}
\section{Experiments on Synthetic Data}
\label{sec:synthetic}
\vspace{-1mm}
\begin{figure*}
     \centering
      \subfloat[Outcome monotonic additive costs]{
         \centering
         \includegraphics[scale=0.3]{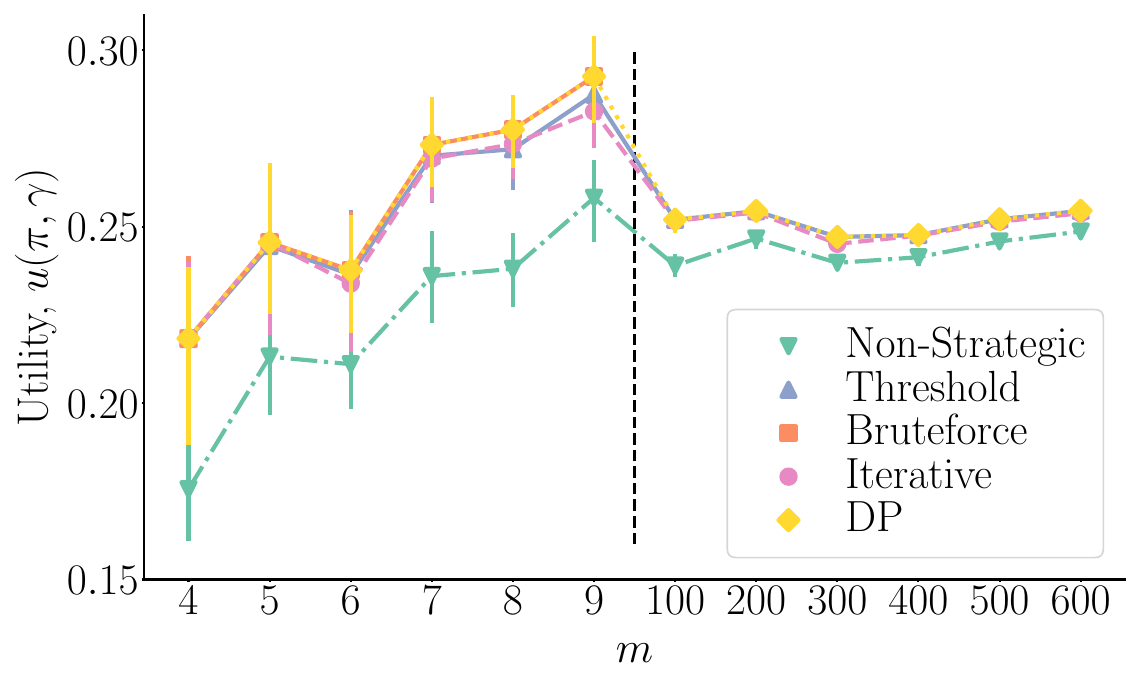}
         \label{fig:synthetic-outcome-monotonic}
     }
     \hspace{15mm}     
     \subfloat[General costs]{
         \centering
         \includegraphics[scale=0.3]{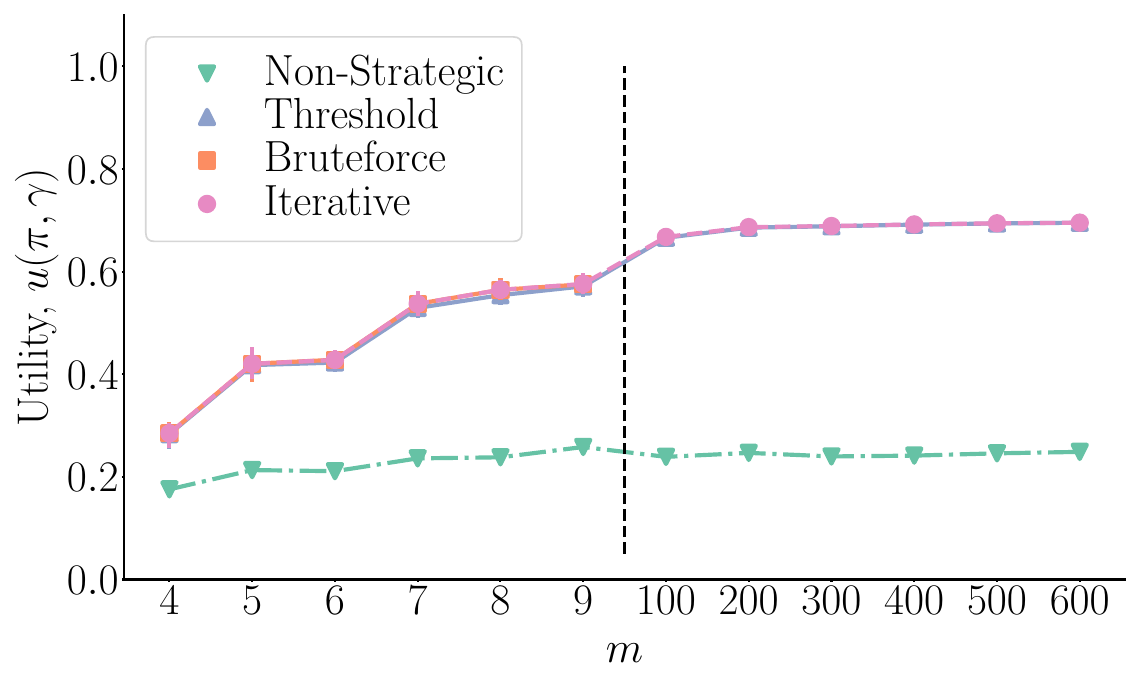}
         \label{fig:synthetic-general}
     }     
     \caption{Performance evaluation on synthetic data. 
     Panels show the utility obtained by several decision policies against the number of feature values $m$.
     %
     Here, note that the dynamic programming (DP) algorithm (Algorithm~\ref{alg:dp}) only works with outcome monotonic additive costs and thus
     only appears in Panel (a). 
     In Panel (a), we set $\kappa=0.1$ and, in Panel (b), we set $\kappa = 0.75$.}
     %
     \label{fig:synthetic}
\end{figure*}
In this section, we evaluate both our dynamic programming algorithm (Algorithm~\ref{alg:dp}) and our iterative algorithm 
(Algorithm~\ref{alg:iterative}) on outcome monotonic and general costs. 
%
We first compare the utility achieved by the decision policies found by our algorithms and those found by several competitive baselines.
%
Then, we compare their computational complexity both in terms of running time and number of rounds (or iterations) to termination. 

\xhdr{Performance evaluation}
We compare the utility achieved by the decision policies found by our algorithms and those found by several baselines. More 
specifically, we consider: 
%
\begin{itemize}[leftmargin=0.7cm] 
\item[(i)] the optimal deterministic threshold rule in a non-strategic setting (Eq.~\ref{eq:detthresh}) [\emph{Non-Strategic}];
\item[(ii)] the optimal deterministic threshold rule in a strategic setting, found via bruteforce search over all deterministic 
threshold rules [\emph{Threshold}];
\item[(iii)] the optimal (stochastic) decision policy in a strategic setting, found via brute force search [\emph{Bruteforce}];
%

%
\item[(iv)] the (stochastic) decision policy found by our dynamic programming algorithm (Algorithm~\ref{alg:dp}), which we can only run
for instances with outcome monotonic additive costs [\emph{DP}].
\item[(v)] the (stochastic) decision policy found by our iterative algorithm (Algorithm~\ref{alg:iterative}) [\emph{Iterative}];
%
%
%
%
\end{itemize}
Here, for simplicity, we consider unidimensional features with $m$ discrete values $\xb \in \{0, \ldots, m-1\}$ and compute $P(\xb=i) = p_i / \sum_j p_j$, where $p_i$
is sampled from a Gaussian distribution $N(\mu=0.5, \sigma=0.1)$ truncated from below at zero.
Then, we sample $P(y=1 \given \xb) \sim U[0,1]$ and we set $\gamma=0.3$.

For instances with outcome monotonic additive costs, we initially set $c(\xb_i,\xb_j)=0\ \forall \xb_i,\xb_j : P(y\given\xb_j)\leq P(y\given\xb_i)$.
Then, we take $m-1$ samples from $U[0,1/\kappa]$ and assign them to $c(\xb_m,\xb_i)\ \forall i<m$ such that $c(\xb_m,\xb_i)>c(\xb_m,\xb_j)\ \forall i<j$ and $\kappa\in(0,1]$.
Finally, we set the remaining values $c(\xb_i,\xb_j)$, in decreasing order of $i$ and $j$ such that $c(\xb_i,\xb_j)=c(\xb_{i-1},\xb_j)-c(\xb_{i-1},\xb_i)$.
It is easy to observe that, proceeding this way, individuals with feature values $\xb_i$ can move (on expectation) to at most $\kappa m$ \emph{better} states, 
\ie,  $c(\xb_i,\xb_j)\leq 1\ \forall \xb_i,\xb_j :\max(1,i-\kappa m)\leq j<i$.
For instances with general costs, we sample the cost between feature values $c(\xb_i, \xb_j) \sim U[0, 1]$ for a fraction $\kappa$ of all pairs and
set $c(\xb_i, \xb_j) = \infty$ for the remaining pairs.
\begin{figure*}[t]
     \centering
     \subfloat[Running time vs. \# feature values]{
         \centering
         \includegraphics[scale=0.25]{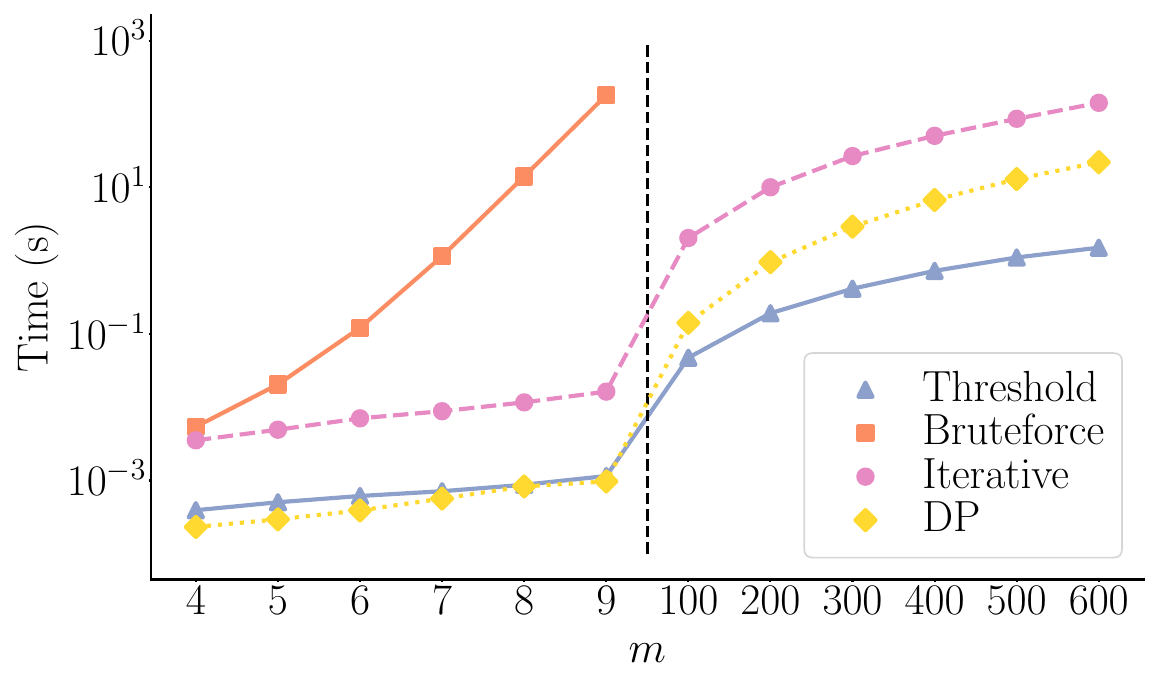}
         \label{fig:runtime-a}
     } \hspace{2mm}
     \subfloat[\# iterations vs. \# feature values]{
         \centering
         \includegraphics[scale=0.25]{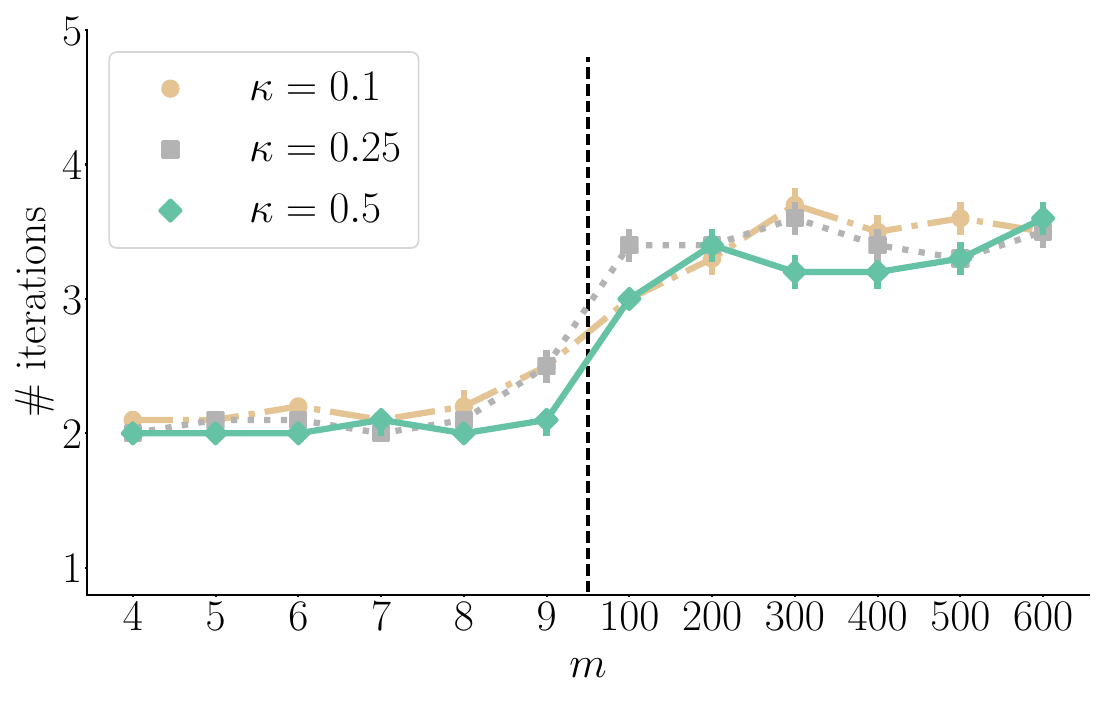}
         \label{fig:runtime-b}
     }
     \subfloat[\# rounds vs. \# feature values]{
         \centering
         \includegraphics[scale=0.25]{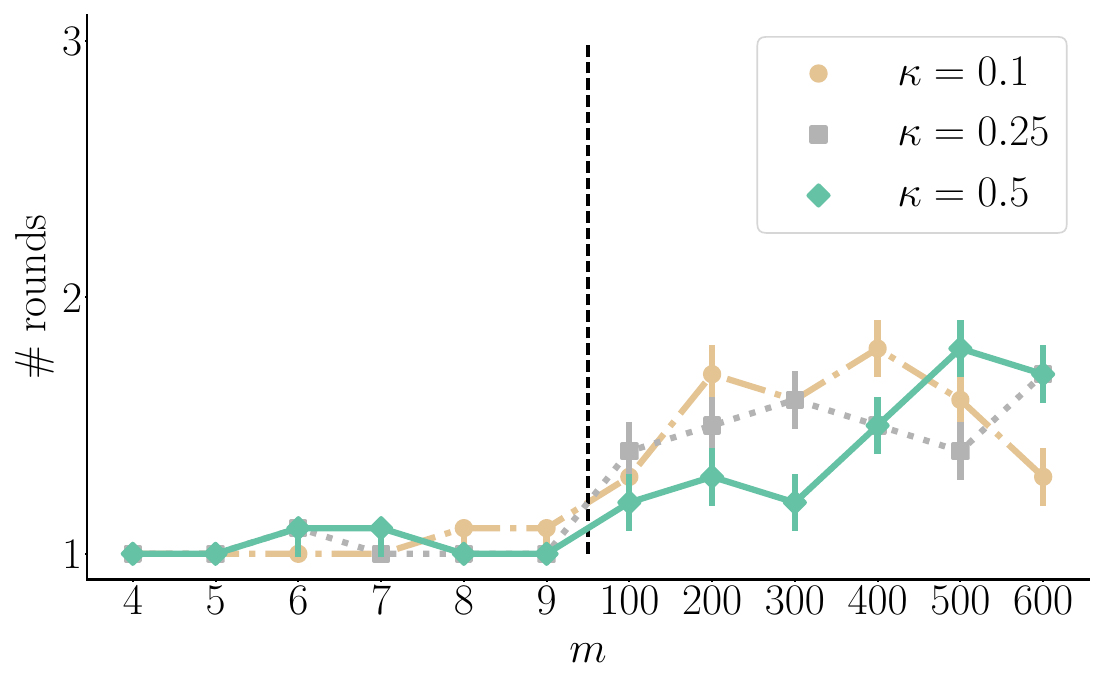}
         \label{fig:runtime-c}
     } \hspace{2mm}
     \caption{Running time analysis on synthetic data with outcome monotonic and additive costs. 
     Panel (a) shows the running time of the brute force search, the threshold policy baseline, our iterative algorithm and our dynamic programming algorithm.
     Panels (b) and (c) show the number of iterations and rounds required by the iterative and dynamic programming algorithms until termination, respectively, 
     for different $\kappa$ values.
     In Panel (a), we set $\kappa=0.1$.
     %
   	}
     \label{fig:runtime}
\end{figure*}

Figure~\ref{fig:synthetic} summarizes the results for both outcome monotonic and general costs, where we repeat each experiment $10$ times 
to obtain error bars. In both cases, we observe that the optimal decision policy in a non-strategic setting has an underwhelming performance.
%
For outcome monotonic additive costs, we observe that the policies found using our dynamic programming algorithm and brute force search closely match 
each other in terms of utility and they consistently outperform the policies found by the iterative algorithm.
For general costs, we find that our iterative algorithm and the threshold policy baseline are the top
performers. 
%

\xhdr{Running time and number of iterations/rounds}
To compare the running time of all the aforementioned algorithms\footnote{\scriptsize We ran all experiments 
on a machine equipped with 48 Intel(R) Xeon(R) 3.00GHz CPU cores and 1.2TB memory.}, we consider the same 
configuration as in the performance evaluation with outcome monotonic and additive costs. 
Figure~\ref{fig:runtime-a} summarizes the results, which show several interesting insights.
We find that brute force search quickly becomes computationally intractable.
%
%
%
Moreover, we observe that the dynamic programming algorithm, is significantly faster than the iterative algorithm, making it the most efficient of the proposed algorithms.
To understand why, we compute the number of iterations/rounds the two algorithms take to terminate in Figures~(\ref{fig:runtime-b},\ref{fig:runtime-c}).
Recall that the complexity of one round in the dynamic programming algorithm and one iteration in the iterative algorithm is $\Ocal(m^3)$.
%
The results show that, 
although in theory, the dynamic programming algorithm needs $\Ocal(m)$ rounds to terminate, in practice, it rarely needs more than two rounds.
This is in contrast with the iterative algorithm which might need a larger number of iterations to converge, especially for large values of $m$.
Overall, the above results let us conclude that, under outcome monotonic additive costs, the dynamic programming algorithm is a highly 
effective and efficient heuristic.

%
%
%

%

\vspace{-2mm}
\section{Experiments on Real Data}
\label{sec:real}
\vspace{-1mm}
In this section, we evaluate our iterative algorithm using real credit 
card data.
%
%
Since in our experiments, the cost individuals pay to change features is not always monotonic, we cannot experiment 
with our dynamic programming algorithm.

\xhdr{Experimental setup}
We use the publicly available~\textit{credit} da\-ta\-set~\citep{yeh2009comparisons}, which contains information about a bank'{}s credit card 
payoffs\footnote{We used a preprocessed version of the credit dataset by~\citet{ustun2019actionable}}.
%
%
For each accepted credit card holder, the respective dataset contains various demographic characteristics 
and financial status indicators which serve as features $\xb$ and the current credit payoff status which serves 
as label $y$. 
Among the features, we distinguish both numerical and discrete-valued features as well as actionable (\eg, most recent bill amount) 
and non-actionable (\eg, marital status) features.
%
Refer to Appendix~\ref{app:raw-features} for more details on the specific features we used.

To approximate the conditional distribution $P(y \given \xb)$, we follow the same procedure as in Tsirtsis and Gomez-Rodriguez~\cite{tsirtsis2020decisions},
which we describe next for completeness.
%
%
%
First, we cluster the credit card holders into $k$ groups based on the original numerical features using $k$-means clustering 
and then, for each credit card holder, we replace their initial numerical features with the respective identifier of the cluster they
belong to, represented using a one-hot encoding.
After this preprocessing step, the discrete feature values $\xb_i$ consist of all possible value combinations of discrete non-actionable features, if 
any, and cluster identifiers.
Then, we train four types of classifiers (Multi-layer perceptron, support vector machine, logistic regression, decision tree) using scikit-learn~\cite{sklearn_api} with default parameters. Finally, we choose the pair of classifier type and number of clusters $k$ that maximizes accuracy, estimated using 5-fold cross validation,
to approximate the values of $P(y \given \xb)$.

To set the cost function $c(\xb_i, \xb_j)$ values, we use the \textit{maximum percentile shift}~\citep{ustun2019actionable}.
More specifically, let $\Lcal$ be the set of actionable (numerical) features and $\bar{\Lcal}$ be the set of non-actionable (discrete-valued) features.
%
%
%
%
Then, for each pair of feature values, we set the cost function $c(\xb_i, \xb_j)$ to:
\begin{equation}\label{eq:cost}
c(\xb_i, \xb_j) = 
\begin{cases}
	\alpha \cdot \max_{l\in\Lcal}|Q_l(x_{j,l})-Q_l(x_{i,l})| & \text{if } x_{i,l}=x_{j,l}\ \forall l\in\bar{\Lcal} \\
    \infty & \text{otherwise},
\end{cases} 
\end{equation}
where $x_{j,l}$ is the value of the $l$-th feature for the feature value $\xb_j$, $Q_l(\cdot)$ is the CDF of the numerical feature $l \in\Lcal$ 
and $\alpha\geq 1$ is a scaling factor which controls the difficulty of changing features.
As an exception, we always set the cost $c(\xb_i, \xb_j)$ between two feature values to $\infty$ if $Q_l(x_{j,l})<Q_l(x_{i,l})$ for $l\in\{\text{Total overdue counts}$, $\text{Total months overdue}\}$, not allowing the history of overdue payments to be erased.
%

Finally, we set the parameter $\gamma$ to the $50$-th percentile of all the individuals'{} $P(y = 1 | x)$, such that $50\%$ of the population is accepted 
by the optimal threshold policy in the non strategic setting.
Refer to Appendix~\ref{app:overview} for further details on the experimental setup.

\begin{figure*}[t]
     \centering
     \subfloat[Non-strategic setting]{
         \centering
         \includegraphics[scale=0.22]{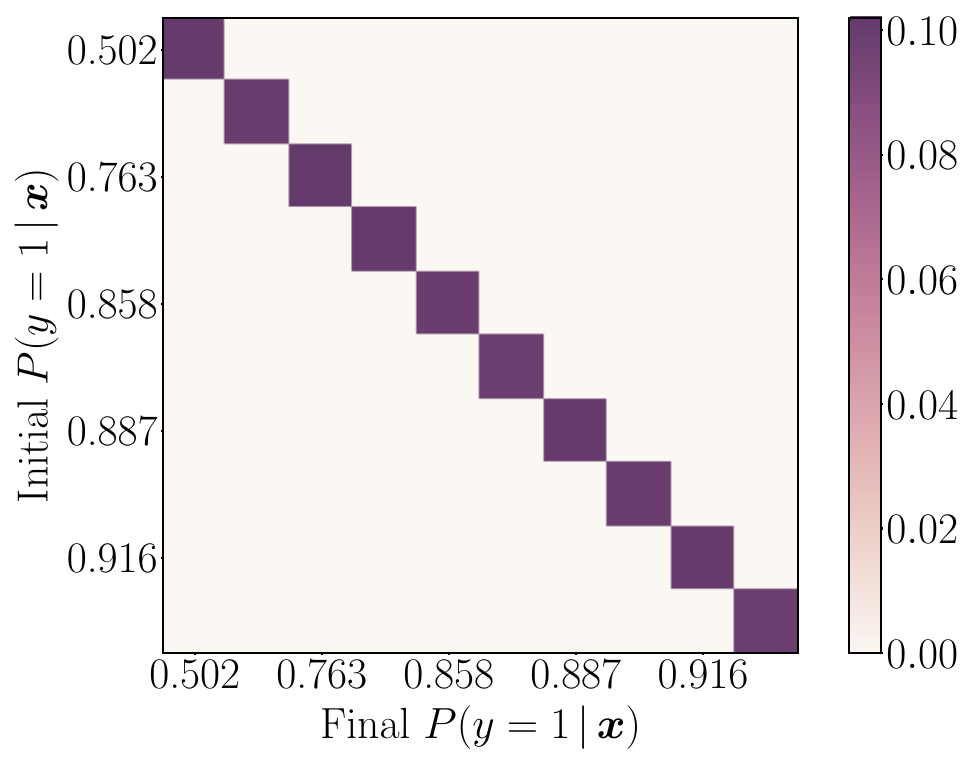}
     } \hspace{2mm}
     \subfloat[\parbox{22mm}{Strategic setting with $\alpha=10$}]{
         \centering
         \includegraphics[scale=0.22]{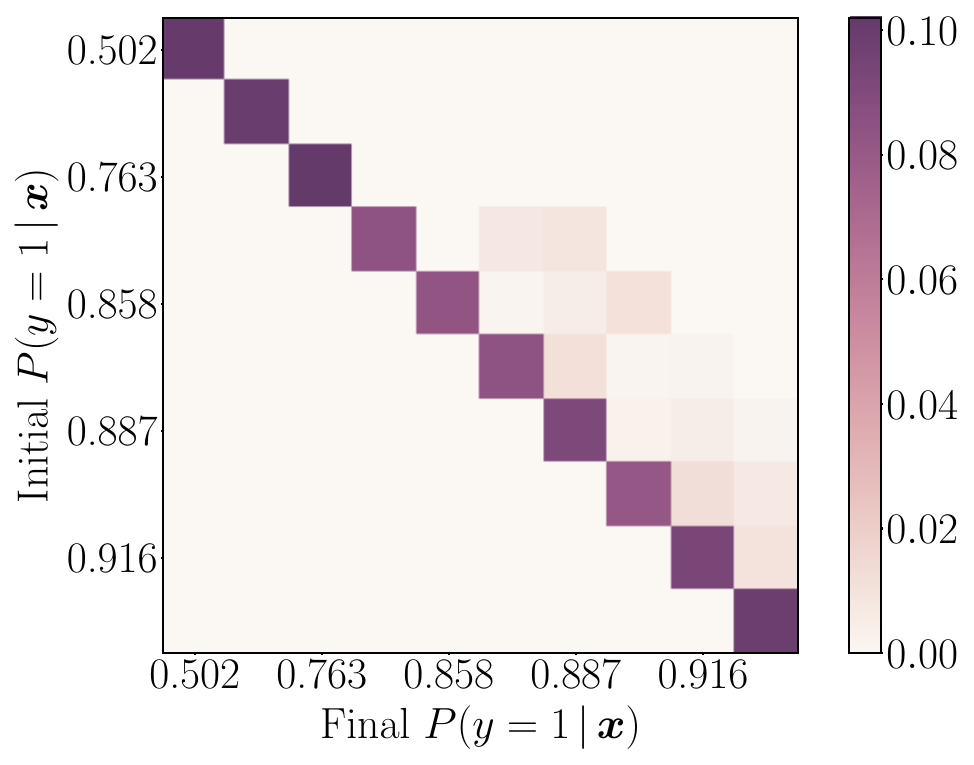}
     } \hspace{2mm}
     \subfloat[\parbox{22mm}{Strategic setting with $\alpha=3.3$}]{
         \centering
         \includegraphics[scale=0.22]{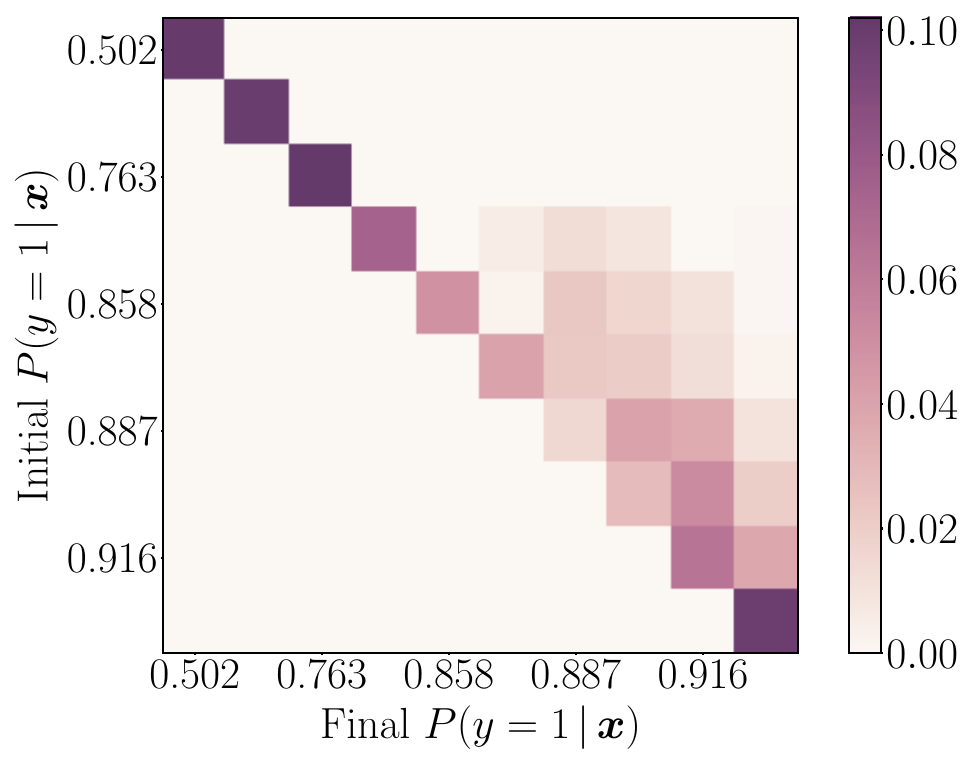}
     } \hspace{2mm}
      \subfloat[\parbox{22mm}{Strategic setting with $\alpha=1$}]{
         \centering
         \includegraphics[scale=0.22]{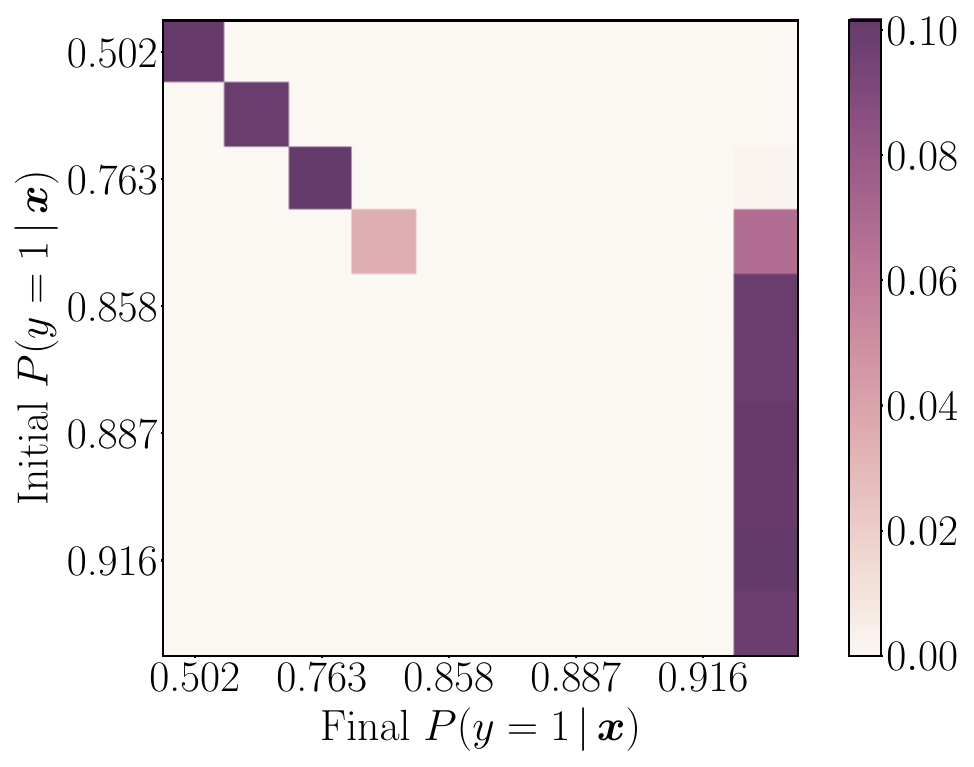}
     }
     \vspace{-1mm}
     \caption{
	Transportation of mass induced by the policies found via the iterative algorithm (Algorithm~\ref{alg:iterative}) in the credit dataset for several values of $\alpha$, which controls the difficulty of changing features.
     For each individual in the population, we record her outcome $P(y=1\given\xb)$ before the best-response (Initial $P(y=1\given\xb)$) and after the best response (Final $P(y=1\given\xb)$).
     In each panel, the color illustrates the percentage of individuals with the corresponding initial and final $P(y=1\given\xb)$ values.
     }
     \label{fig:transportation}
\end{figure*}

\begin{figure*}[t]
     \centering
     \subfloat[Utility vs. $1/\alpha$ ]{\label{fig:real_utility}
         \centering
         \includegraphics[scale=0.25]{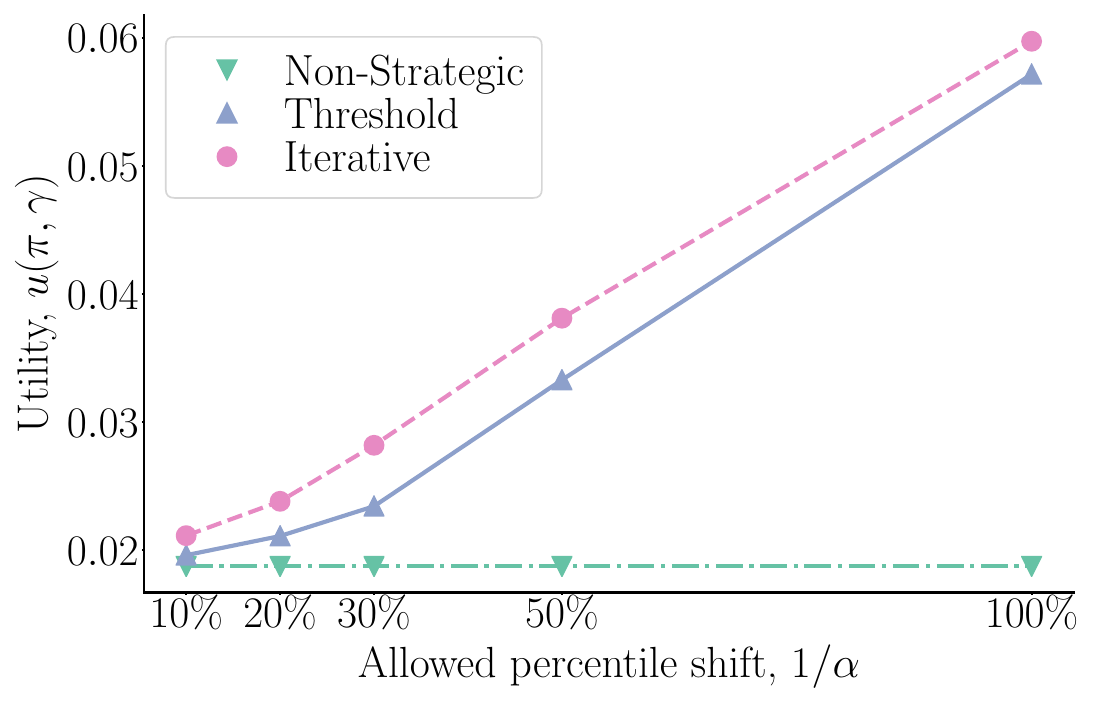}
     } \hspace{2mm}
     \subfloat[Running time vs. $1/\alpha$ ]{\label{fig:real_time}
         \centering
         \includegraphics[scale=0.25]{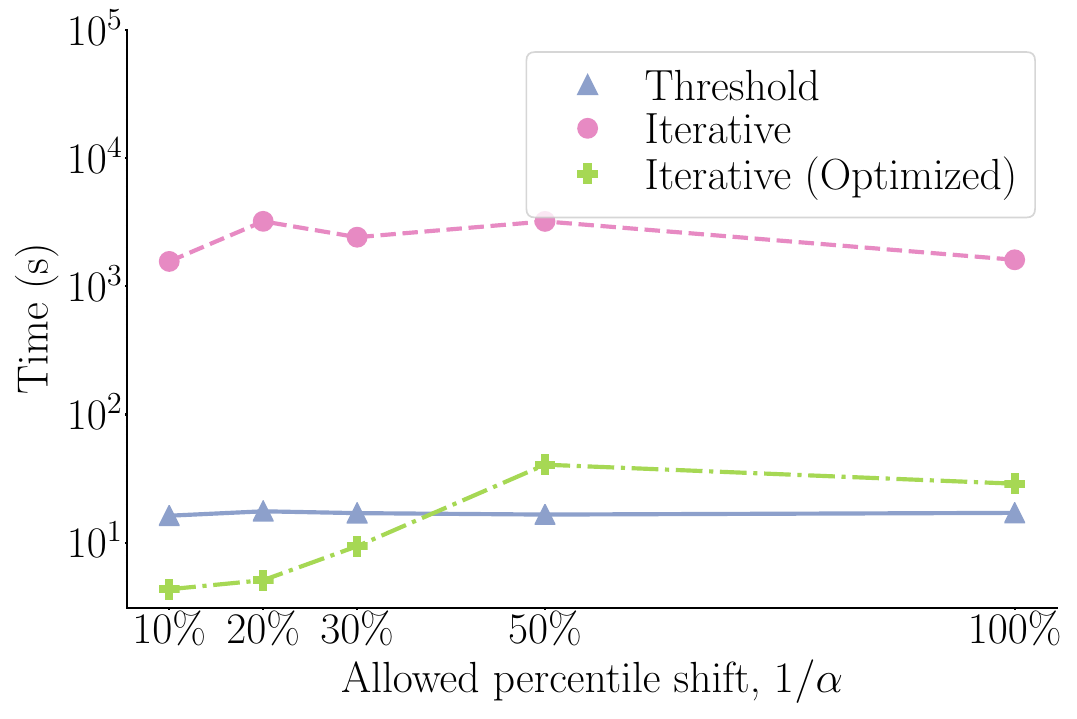}
     } \hspace{2mm}
      \subfloat[\# components vs. $1/\alpha$ ]{\label{fig:real_components}
         \centering
         \includegraphics[scale=0.25]{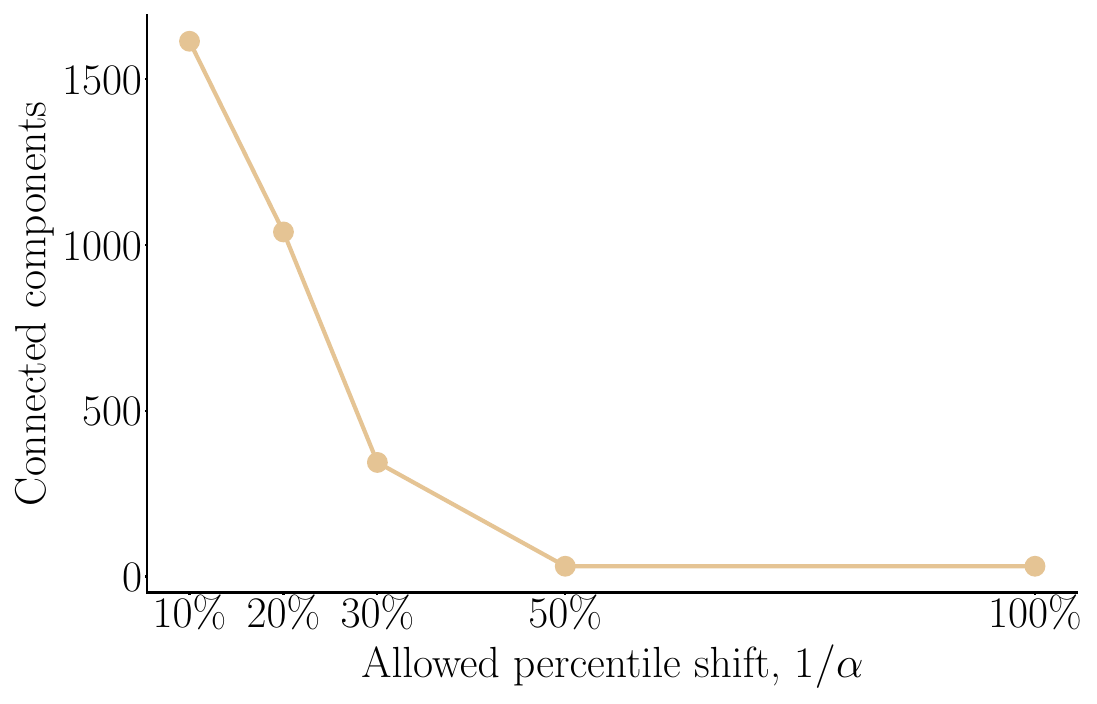}
     }
     \caption{Effectiveness and efficiency of the proposed algorithms.
     Panel (a) shows the utility achieved by three types of decision policies in the credit dataset, against the value of the parameter $\alpha$, which controls how difficult it is for the individuals to change their features.
     Panel (b) shows the running time of the threshold policy baseline algorithm and our iterative algorithm, with and without the speed-up discussed in Section~\ref{sec:general}.
	Panel (c) shows the number of connected components in the graph $\Gcal$.
	%
	%
	Note that, whenever we implement the iterative algorithm with the speed up, we solve the subproblems corresponding to independent components sequentially, 
	however, the procedure is amenable to parallelization. 
     }
     \label{fig:analysis}
\end{figure*}

\xhdr{Results}
We first look into the transportation of mass induced by the decision policy found by our iterative algorithm for different
$\alpha$ values in Figure~\ref{fig:transportation}.
We observe that, as the cost of changing features increases, there is a higher transportation of mass towards feature values
with the highest outcomes $P(y=1\given\xb)$.
Moreover, whenever individuals can arbitrarily change actionable features (\ie, $\alpha=1$), the best-response of individuals
is either feature values with the highest outcomes or their initial features if their recourse may be limited due to non-actionable
features (\eg, history of overdue payments).

Next, we compare the utility of the decision policy found by our iterative algorithm and the policies found by the same baselines used in 
Section~\ref{sec:synthetic}.
Here, we do not compare with the optimal (stochastic) decision policy because brute force search does not 
scale to the size of the dataset.
Figure~\ref{fig:real_utility} summarizes the results for several values of the cost scaling factor $\alpha$, which
show that the decision policy found by the iterative algorithm outperforms the baselines and, as the cost of changing features becomes smaller ($\alpha$ decreases), the utility value increases.

Finally, we compare the running time of the threshold baseline and the iterative algorithm with and without the speed up that exploits the presence
of non-actionable features, described in Section~\ref{sec:general}.
Figures~\ref{fig:real_time},~\ref{fig:real_components} summarize the results. We observe that, whenever the cost to change features is high, there exist many independent 
connected components and the speed up provides a significant advantage. 
In those cases, the iterative algorithm with the speed up performs faster than the threshold baseline while the running time of the two algorithms remains comparable, even when the cost to change features is low. 
%
%
%

\vspace{-2mm}
\section{Conclusions}
\label{sec:conclusions}
\vspace{-1mm}
In this paper, we have studied the problem of finding optimal decision policies that maximize utility in a strategic
setting.
We have shown that, in contrast with the non-strategic setting, optimal decision policies that maximize utility are hard 
to find.
However, if the cost individuals pay to change their features satisfies a natural monotonicity assumption, we have 
demonstrated that we can narrow down the search for the optimal policy to a particular family of decision policies,
which allow for a highly effective polynomial time dynamic programming heuristic search algorithm.
Finally, we have lifted the monotonicity assumption and developed an efficient iterative search algorithm that is guaranteed
to find locally optimal decision policies also in polynomial time.

Our work opens up many interesting avenues for future work.
For example, in our experiments, we have found that our dynamic programming algorithm often finds outcome monotonic binary 
policies with optimal utility. However, we are lacking theoretical guarantees, \eg, approximation guarantees, supporting this good 
empirical performance. 
%
%
Throughout the paper, we have assumed that features take discrete values. It would be very interesting to extend our work to real valued
features.
Moreover, we have considered policies that maximize utility. A natural step would be considering utility maximization under
fairness constraints~\cite{hardt2016equality, zafar2017fairness, zafar2019fairness}. 
%
Similarly as in previous work in strategic classification, the individuals have white-box access to the decision policy, however, in practice, they 
may only have access to \emph{explanations} of specific outcomes.
Finally, there are reasons to believe that causal features should be more robust to strategic behavior~\cite{miller2019strategic}. It would 
be interesting to investigate the use of causally aware feature selection methods~\cite{rojas2018invariant} in strategic settings. 
%

{
\bibliographystyle{plainnat}
\bibliography{consequential-decisions}
}

\clearpage
\newpage
\onecolumn

\appendix
\section{Proofs}

\subsection{Proof of Theorem~\ref{thm:hardness}} \label{app:hardness}
We will reduce any given instance of the SAT problem~\citep{karp1972reducibility}, which is known to be NP-complete, to a particular instance of our problem. 
In a SAT problem, the goal is finding the value of a set of boolean variables $\{v_1, v_2, \dots, v_l\}$, and their logical complements $\{\bar{v}_1, \bar{v}_2, \dots, \bar{v}_l\}$, 
that satisfy $s$ number of \textit{OR} clauses, which we label as $\{k_1, k_2, \dots, k_s\}$.

First, we start by representing our problem, as defined in Eq.~\ref{eq:utility-maximization-2}, using a directed weighted bipartite graph, whose nodes can be divided into two disjoint 
sets $U$ and $V$. In each of these sets, there are $m$ nodes with labels $\{x_1, \ldots, x_m\}$.
We characterize each node $x_i$ in $U$ with $P(x_i)$ and each node $x_j$ in $V$ with $\pi(x_j)$ and $u(x_j) = P(y = 1 \given x_j) - \gamma$. Then, we connect each node $x_i$ 
in $U$ to each node $x_j$ in $V$ and characterize each edge with a weight $w(x_i, x_j) = b(x_j) - c(x_i, x_j) = \pi(x_j) - c(x_i, x_j)$ and a utility
\begin{equation*}
u(x_i, x_j) = \pi(x_j) u(x_j) P(x_i) \II(x_j = \argmax_{x_k} w(x_i, x_k)),
\end{equation*}
which, for each node $x_i$ in $U$, is only nonzero for the edge with maximum weight (solving ties at random). 
Under this representation, the problem reduces to finding the values of $\pi(x_j)$ such that the sum of the utilities of all edges in the graph is maximized.

Next, given an instance of the SAT problem with $\{v_1, v_2, \dots, v_l\}$ and $\{k_1, k_2, \dots, k_s\}$, we use the above representation to build the following instance of
our problem. More specifically, consider $U$ and $V$ have $m = 7l + s$ nodes each with labels
\begin{equation*}
\{ y_1, \dots, y_l, \bar{y}_1, \dots, \bar{y}_l, a_1, \dots, a_l, b_1, \dots, b_l, z_{11}, \dots, z_{1l}, z_{21}, \dots, z_{2l}, z_{31}, \dots, z_{3l}, k_1, k_2, \dots, k_s\}
\end{equation*}
For the set $U$, characterize each node $u$ with $P(u)$, where
\begin{align*}
P(z_{1i}) = \frac{3(s+1)}{3l + 3(s+1)l}, \,\, P(z_{2i}) &= P(z_{3i}) = \frac{1}{3l + 3(s+1)l}, \,\, P(k_{j}) = \frac{1}{3l + 3(s+1)l}, \,\, \mbox{and} \\
P(y_i) &= P(\bar{y}_i) = P(a_i) = P(b_i) = 0,
\end{align*}
for all $i=1, \ldots, l$ and $j=1, \ldots, s$.
For the set $V$, characterize each node $v$ with $\pi(v)$ and $u(v)$, where
\begin{equation*}
u(y_i) = u(\bar{y}_1) = \frac{1}{2l+4(s+1)l}, \,\, u(a_i) = u(b_i) = \frac{2(s+1)}{2l+4(s+1)l}, \,\, u(z_{1i}) = u(z_{2i}) = u(z_{3i}) = 0, \,\, \text{and} \,\, u(k_{j}) = 0, 
\end{equation*}
for all $i=1, \ldots, l$ and $j=1, \ldots, s$.
%
Then, connect each node $u$ in $U$ to each node $v$ in $V$ and set each edge weights to $w(u, v) = \pi(v) - c(u, v)$, where:
\begin{itemize}[noitemsep,nolistsep,leftmargin=0.8cm]
\item[(i)] $c(z_{1i}, y_j) = c(z_{1i}, \bar{y}_j) = 0$ and $c(z_{2i}, y_j) = c(z_{3i}, y_j) = c(k_{q}, y_j) = c(z_{2i}, \bar{y}_j) = c(z_{3i}, \bar{y}_j) = c(k_{q}, \bar{y}_j) = 2$ for each $i, j=1, \ldots, l$ and $q=1, \ldots, s$.
\item[(ii)] $c(z_{2i}, y_j) = 0$, $c(z_{2i}, a_j) = 1-\epsilon$ and $c(z_{1i}, y_j) = c(z_{3i}, y_j) = c(k_{q}, y_j) = c(z_{1i}, a_j) = c(z_{3i}, a_j) = c(k_{q}, a_j) = 2$ for each $i, j=1, \ldots, l$ and $q=1, \ldots, s$.
\item[(iii)] $c(z_{3i}, \bar{y}_j) = 0$, $c(z_{3i}, b_j) = 1-\epsilon$ and $c(z_{1i}, \bar{y}_j) = c(z_{2i}, \bar{y}_j) = c(k_{q}, \bar{y}_j) = c(z_{1i}, b_j) = c(z_{2i}, b_j) = c(k_{q}, b_j) = 2$ for each $i, j=1, \ldots, l$ and $q=1, \ldots, s$.
\item[(iv)] $c(k_i, y_j) = 0$ if the clause $k_i$ contains $y_j$, $c(k_i, \bar{y}_j) = 0$ if the clause $k_i$ contains $\bar{y}_j$, and $c(k_i, a_j) = c(k_i, b_j) = 2$ for all $i=1, \ldots, s$ and $j=1, \ldots, l$.
\item[(v)] All remaining edge weights, set $c(\cdot, \cdot) = \infty$.
\end{itemize}
%
%
%
Now, note that, in this particular instance, finding the optimal values of $\pi(v)$ such that the sum of the utilities of all edges in the graph is maximized reduces to first
solving $l$ independent problems, one per pair $y_j$ and $\bar{y}_j$, since whenever $c(u, v) = 2$, the edge will never be active, and each optimal $\pi$ value will be always either
zero or one.
Moreover, the maximum utility due to the nodes $k_z$ will be always smaller than the utility due to $y_j$ and $\bar{y}_j$ and we can exclude them by the moment.
In the following, we fix $j$ and compute the sum of utilities for all possible values of $y_j$ and $\bar{y}_j$:
\begin{itemize}[noitemsep,nolistsep,leftmargin=0.8cm]
    \item For $\pi(y_j) = \pi(\bar{y}_j) = 0$, the maximum sum of utilities is $\frac{4(s+1)}{(3l + 3(s+1)l)\cdot(2l+4(s+1)l)}$ whenever $\pi(a_j) = \pi(b_j) =1$.
    \item For $\pi(y_j) = \pi(\bar{y}_j) = 1$, the sum of utilities is $\frac{3(s+1)+2}{(3l + 3(s+1)l)\cdot(2l+4(s+1)l)}$ for any value of $\pi(a_j)$ and $\pi(b_j)$.

    \item For $\pi(y_j) = 1-\pi(\bar{y}_j)$, the maximum sum of utilities is $\frac{5(s+1)}{(3l + 3(s+1)l)\cdot(2l+4(s+1)l)}$.
\end{itemize}
Therefore, the maximum sum of utilities $\frac{5(s+1)}{(3 + 3(s+1))\cdot(2l+4(s+1)l)}$ occurs whenever $\pi(y_j) = 1-\pi(\bar{y}_j)$ for all $j = 1, \ldots, l$
and the solution that maximizes the overall utility, including the utility due to the nodes $k_z$, gives us the solution of the SAT problem.
This concludes the proof.

\subsection{Proof of Proposition~\ref{prop:outcome-monotonicity}} \label{app:outcome-monotonicity}
This proposition can be easily proven by contradiction. 
More specifically, assume that all outcome monotonic policies $\pi$ are suboptimal, \ie, $u(\pi, \gamma) < u(\pi^{*}, \gamma)$, where $\pi^{*}$ is an optimal policy
that maximizes utility, and sort the values of the optimal policy in decreasing order, \ie, $1 = \pi^*(\xb_{l_1})\geq \pi^*(\xb_{l_2})\geq ... \geq \pi^*(\xb_{l_n})$. 
Here, note that there is a state $\xb_k$ such that $l_k \neq k$, otherwise, the policy $\pi^{*}$ would be outcome monotonic.
Now, define the index $r=\argmin_k l_k\neq k$ and build a policy $\pi'$ such that $\pi'(\xb_l) = \pi^*(\xb_{l_r})$ for all $r-1 < l < l_r$ and 
$\pi'(\xb_l)=\pi^*(\xb_l)$ otherwise.
Then, it is easy to see that the policy $\pi'$ has greater or equal utility than $\pi^{*}$ and it holds that 
$\pi'(\xb_l) \geq \pi'(\xb_t) \Leftrightarrow P(y\given \xb_l) \geq P(y\given \xb_t)$ for all $\xb_l , \xb_t$ such that 
$l, t \leq l_r$.
If the policy $\pi'$ satisfies outcome monotonicity, we are done. Otherwise, we repeat the procedure starting from $\pi'$
and continue building increasingly better policies until we eventually build one that satisfies outcome monotonicity.
By construction, this last policy will achieve equal or greater utility than the policy $\pi^{*}$, leading to a contradiction.

\subsection{Proof of Theorem~\ref{thm:outcome-monotonic-binary-policy}} \label{app:outcome-monotonic-binary-policy}
We prove this theorem by contradiction. More specifically, assume that all outcome monotonic binary policies $\pi$ are suboptimal, \ie, $u(\pi, \gamma) < u(\pi^{*}, \gamma)$,
where $\pi^{*}$ is an optimal policy.
According to Proposition~\ref{prop:outcome-monotonicity}, under outcome monotonic costs, there is always an optimal outcome monotonic 
policy. Now, assume there is an optimal outcome monotonic policy $\pi^{*}$ such that $\pi^*(\xb_{i-1})>\pi^*(\xb_i)>\pi^*(\xb_{i-1})-c(\xb_i,\xb_{i-1})
\vee \pi^*(\xb_i)<\pi^*(\xb_{i-1})-c(\xb_i,\xb_{i-1})$ for some $i > 1$. Moreover, if there are more than one $i$, consider the one with the highest outcome
$P(y\given \xb_i)$. Then, we analyze each case separately.

If $\pi^*(\xb_{i-1})>\pi^*(\xb_i)>\pi^*(\xb_{i-1})-c(\xb_i, \xb_{i-1})$, we can show that the policy $\pi'$ with $\pi'(\xb_j)=\pi^*(\xb_j)\ \forall j\neq i$ and
$\pi'(\xb_i)=\pi^*(\xb_{i-1})$ has greater or equal utility than $\pi^{*}$. 
More specifically, consider an individual with initial feature values $\xb_k$. Then, it is easy to see that, if $k < i$, the best response under $\pi^{*}$ 
and $\pi'$ will be the same and, if $k \geq i$, the best response will be either the same or change to $\xb_i$ under $\pi'$. In the latter case, it also 
holds that $P(y \given x_i) > P(y \given x_j)$, where $x_j$ is the best response under $\pi^{*}$, otherwise, we would have a contradiction. Therefore, 
we can conclude that $\pi'$ provides higher utility than $\pi^{*}$.

If $\pi^*(\xb_i)<\pi^*(\xb_{i-1})-c(\xb_i, \xb_{i-1})$, we can show that the policy $\pi'$ with $\pi'(\xb_j)=\pi^*(\xb_j)\ \forall j\neq i$ and 
$\pi'(\xb_i)=\pi^*(\xb_{i-1})-c(\xb_i,\xb_{i-1})$ has greater or equal utility than $\pi^{*}$.
More specifically, consider an individual with initial feature values $\xb_k$ and denote the individual'{}s best response under $\pi^{*}$ as $\xb_j$.
Then, it is easy to see that the individual'{}s best response is the same under $\pi^{*}$ and $\pi'$, however, if $\xb_j = \xb_i$, the term in the utility 
corresponding to the individual does increase under $\pi'$. Therefore, we can conclude that $\pi'$ provides higher utility than $\pi^{*}$.

In both cases, if the policy $\pi'$ is an outcome monotonic binary policy, we are done, otherwise, we repeat the procedure starting from the corresponding
$\pi'$ and continue building increasingly better policies until we eventually build one that is an outcome monotonic binary policy. By construction, this last
policy will achieve equal or greater utility than the policy $\pi^{*}$, leading to a contradiction.

\subsection{Proof of Proposition~\ref{prop:best-response}} \label{app:best-response}

Consider an individual with initial features $\xb_i$ such that $P(y\given\xb_i)>\gamma$. As argued just after Proposition~\ref{prop:outcome-monotonicity}, given an individual with a set of features $\xb_i$, any outcome monotonic policy always 
induces a best response $\xb_l$ such that $P(y \given \xb_l) \geq P(y \given \xb_i)$, that means, $l < i$. 
Then, we just need to prove that the best response $\xb_l$ cannot satisfy that $P(y\given \xb_l) > P(y \given \xb_j)$ nor satisfy that 
$P(y\given \xb_j) > P(y\given \xb_l) \geq P(y\given \xb_i)$, where $j = \max \{k \given k \leq i, \pi(\xb_k) = 1 \vee \pi(\xb_k) = \pi(\xb_{k-1}) \}$.
Without loss of generality, we assume that $j<i$, however, in case $j=i$ the main idea of the proof is the same.

First, assume that $P(y\given \xb_l) > P(y\given \xb_j)$. Then, using the additivity and outcome monotonicity of the cost and the fact that the policy is an outcome
monotonic binary policy, it should hold that 
$\pi(\xb_j) - c(\xb_i,\xb_j) = \pi(\xb_{j-1})-c(\xb_i, \xb_j) > \pi(\xb_{j-1}) - c(\xb_i,\xb_{j-1}) \geq \pi(\xb_{j-2}) - c(\xb_i, \xb_{j-2}) \geq \dots \geq \pi(\xb_l)-c(\xb_i,\xb_l)$. 
This implies that $\xb_j$ is a strictly better response for the individual than $\xb_l$, which is a contradiction.
Now, assume that $P(y\given \xb_j) > P(y\given \xb_l) \geq P(y\given \xb_i)$. Then, using the additivity of the cost, the definition of $\xb_j$ and the fact that $\xb_l$ is the 
best-response, it should hold that $\pi(\xb_j) - c(\xb_i, \xb_j) < \pi(\xb_l) - c(\xb_i,\xb_l) = \pi(\xb_j) - c(\xb_l,\xb_j) - c(\xb_i, \xb_l) = \pi(\xb_j)-c(\xb_i,\xb_j)$, which is clearly 
a contradiction.
Therefore, $\xb_j$ is a best-response.

Now, consider an individual with initial features $\xb_i$ such that $P(y\given\xb_i)\leq\gamma$ and $\pi(\xb_j)\geq c(\xb_i,\xb_j)$.
The argument for proving that $P(y\given \xb_l)>P(y\given \xb_j)$ is a contradiction remains as is.
Assume that $P(y\given \xb_j) > P(y\given \xb_l) \geq P(y\given \xb_i)$.
Then $\pi(\xb_l)=\pi(\xb_j)-c(\xb_l,\xb_j)$ or $\pi(\xb_l)=0$, meaning that $\pi(\xb_j)-c(\xb_l,\xb_j)>\pi(\xb_l)$ since $\pi(\xb_j)-c(\xb_l,\xb_j)>\pi(\xb_j)-c(\xb_i,\xb_j)\geq 0$.
Therefore, it should hold that $\pi(\xb_j) - c(\xb_i, \xb_j) < \pi(\xb_l) - c(\xb_i,\xb_l) \leq \pi(\xb_j) - c(\xb_l,\xb_j) - c(\xb_i, \xb_l) = \pi(\xb_j)-c(\xb_i,\xb_j)$, which is clearly 
a contradiction.
As a result, $\xb_j$ is a best-response.

Now, consider an individual with initial features $\xb_i$ such that $P(y\given\xb_i)\leq\gamma$ and $\pi(\xb_j)< c(\xb_i,\xb_j)$.
The argument for proving that $P(y\given \xb_l)>P(y\given \xb_j)$ is a contradiction remains as is.
For all $\xb_l$ such that $P(y\given \xb_j) \geq P(y\given \xb_l) > P(y\given \xb_i)$ we have $\pi(\xb_l)=\pi(\xb_j)-c(\xb_l,\xb_j)$ meaning that $\pi(\xb_l)-c(\xb_i,\xb_l)=\pi(\xb_j)-c(\xb_i,\xb_j)<0$ or $\pi(\xb_l)=0$ meaning that $\pi(\xb_l)-c(\xb_i,\xb_l)<0$.
In both cases, because $\pi(\xb_i)=0$, we get that $\xb_i$ is a best-response. 

\subsection{Proof of Proposition~\ref{prop:termination}} \label{app:termination}

We prove that $\bar{u}$ is a denominator of $\pi(\xb_j)$ $\forall \xb_j \in \{1, \ldots, m\}$ after each step in the iterative algorithm.
We prove this claim by induction. The induction basis is obvious as we initialize the values of $\pi(\xb_j) = 0$ for all $\xb_j$. For the induction step, suppose that we are going
to update $\pi(\xb_j)$ in our iterative algorithm. According to the induction hypothesis we know that ${\pi(\xb_k) \over \bar{u}} \in \mathcal{Z}$ $\forall \xb_k \in \{1, \ldots, m\}$.
Then, it can be shown that the new value of $\pi(\xb_j)$ will be chosen among the elements of the following set (these are the thresholds that might change the
transfer of masses):
\begin{equation*}
   \pi_{new}(\xb_j) \in \{0\} \cup \{1\} \cup \left\{ max_k ( \pi(\xb_k)-c(\xb_i,\xb_k) )+ c(\xb_i, \xb_j) \given \xb_i \in \{0, \ldots, m\} \right\}
\end{equation*}
In the above, it is clear that all these possible values are divisible by $\bar{u}$, so the new value of $\pi(\xb_j)$ will be divisible by $\bar{u}$
too.
Then, since $0 \leq \pi(\xb_j) \leq 1$ and ${\pi(\xb_j) \over \bar{u}} \in \mathcal{Z}$ for all $\xb_j \in \{1, \ldots, m\}$, there are $1+{1\over \bar{u}}$
possible values for each $\pi(\xb_j)$, \ie, $0,\bar{u},2\bar{u},\ldots,1$.
As a result, there are $m^{1+{1\over\bar{u}}}$ different decision policies $\pi$.
Finally, since the total utility increases after each step, the decision policy $\pi$ at each step must be different. As a result, the algorithm will terminate 
after at most $m^{1+{1 \over \bar{u}}} -1$ steps.

\clearpage
\newpage

\section{Additional details on the experiments on real data} \label{app:real-data}

\subsection{Raw features} \label{app:raw-features}

 
Each credit card holder has a label which indicates whether they will default during the next month ($y=0$) or not ($y=1$) and the features $\xb$ are:
 \begin{itemize}
    \item Marital status: whether the person is married or single.
    \item Age group: group depending on the person's age ($<25$, $25-39$, $40-59$, $>60$).
    \item Education level: the level of education the individual has acquired (1-4).
    \item Maximum bill amount over last 6 months
    \item Maximum payment amount over last 6 Months
    \item Months with zero balance over last 6 Months
    \item Months with low spending over last 6 Months
    \item Months with high spending over last 6 Months
    \item Most recent bill amount
    \item Most recent payment amount
    \item Total overdue counts
    \item Total months overdue
 \end{itemize}
We consider all features except marital status, age group and education level to be actionable and, among the actionable features, 
we assume that total overdue counts and total months overdue can only increase.

\subsection{Further details on the experimental setup for the credit card dataset} \label{app:overview}
%
%
Table~\ref{tab:data} summarizes the experimental setup for the credit card dataset, \ie, number of samples, the pair of classifier - number 
of clusters $k$ picked through cross-validation, the accuracy achieved by the corresponding classifier, the resulting number of feature values $m$ and 
the parameter $\gamma$.

\begin{table}[t]
\caption{Dataset details }
\begin{center}
\begin{tabular}{ |l||c|c|c|c|c|c|c| } 
 \hline
 Dataset & \# of samples & Classifier & $k$ & Accuracy & $m$ & $\gamma$ \\ \hline 
 credit & $30000$ & Logistic Regression & $100$ & $80.4\%$ & $3200$ & $0.85$ \\ 
 \hline
\end{tabular}
\end{center}
\end{table}\label{tab:data}

%
%
%
%

\end{document}